\documentclass[10pt,twocolumn,letterpaper]{article}

\usepackage{iccv}
\usepackage{times}
\usepackage{epsfig}
\usepackage{graphicx}
\usepackage{amsmath}
\usepackage{amssymb}
\usepackage[utf8]{inputenc}

\usepackage{url}
\usepackage{mathrsfs}
\usepackage[table]{xcolor}
\usepackage{xcolor}
\usepackage{dsfont}
\usepackage{xspace}
\usepackage{listings}
\usepackage{numprint}
\usepackage{multirow}
\usepackage{setspace,lipsum}
\usepackage{booktabs}
\usepackage{microtype}
\usepackage{graphicx}
\usepackage{subfigure}
\usepackage{booktabs} 
\usepackage{balance}

\usepackage{color}
\definecolor{mygreen}{rgb}{0.9,0.65,0.8}
\definecolor{mylightorange}{rgb}{0.9,0.72,0.55}
\definecolor{mycyan}{rgb}{0.7,0.9,1}
\definecolor{mygray}{rgb}{0.92,0.92,0.92}

\usepackage{amsmath, amsthm}
\usepackage{amssymb}
\newcommand{\mcL}{\mathcal{L}}

\newcommand{\fvec}{\mathbf{f}}

\newcommand{\Ccal}{\mathcal{C}}

\newcommand{\Fcal}{\mathcal{F}}

\newcommand{\Xcal}{\mathcal{X}}

\newcommand{\RR}{\mathbb{R}}
\newcommand{\SSD}{\mathbb{S}^{D-1}}

\newcommand{\dsf}{\mathsf{d}}

\newcommand{\muu}{\boldsymbol{\mu}}

 \DeclareMathOperator*{\argmin}{arg\,min}

\usepackage[pagebackref=true,breaklinks=true,letterpaper=true,colorlinks,bookmarks=false]{hyperref}

\iccvfinalcopy 

\def\iccvPaperID{3489} 

\begin{document}

\title{Video Face Clustering with Unknown Number of Clusters
\vspace{-3mm}
}

\author{
Makarand Tapaswi$^{1,2,3}$ \quad~~ Marc T. Law$^{2,3,4}$ \quad~~ Sanja Fidler$^{2,3,4}$ \\
$^1$Inria \quad
$^2$University of Toronto \quad
$^3$Vector Institute \quad
$^4$NVIDIA \\
{\tt\small makarand.tapaswi@inria.fr, \{makarand,law,fidler\}@cs.toronto.edu} \\
{\footnotesize \url{https://github.com/makarandtapaswi/BallClustering\_ICCV2019}}
}

\maketitle

\begin{abstract}
Understanding videos such as TV series and movies requires analyzing who the characters are and what they are doing.
We address the challenging problem of clustering face tracks based on their identity.
Different from previous work in this area, we choose to operate in a realistic and difficult setting where: (i) the number of characters is not known a priori; and (ii) face tracks belonging to minor or background characters are not discarded.

To this end, we propose \emph{Ball Cluster Learning} (BCL), a supervised approach to carve the embedding space into balls of equal size, one for each cluster.
The learned ball radius is easily translated to a stopping criterion for iterative merging algorithms.
This gives BCL the ability to estimate the number of clusters as well as their assignment, achieving promising results on commonly used datasets.
We also present a thorough discussion of how existing metric learning literature can be adapted for this task.
\end{abstract}


\vspace{-5mm}
\section{Introduction}
\label{sec:intro}

Characters are a central aspect of any story.
While video streaming platforms such as Netflix provide the ability to find a movie based on metadata, searching a video collection to find the right clip when ``Jack Sparrow first meets Will'' requires analyzing the content of the video.
Understanding characters also has a direct influence on important research such as video captioning~\cite{rohrbach2017groundedpeople,rohrbach2017lsmdc},  question-answering~\cite{lei2018tvqa,tapaswi2016movieqa}, studying social situations~\cite{vicol2018moviegraphs} and 4D effects~\cite{zhou2018movie4d}.

Characters are often studied by analyzing face tracks (sequences of temporally related detections) in videos.
A significant part of this analysis is \emph{identification} - labeling face tracks with their names, and typically employs supervision from web images~\cite{aljundi2016imdb,nagrani2017sherlock}, transcripts~\cite{baeuml2013semisupervised,everingham2006}, or even dialogs~\cite{cour2010subtitle,haurilet2016naming}.
We are interested in an equally popular alternative - \emph{clustering} face tracks based on identity.
Note that clustering is complementary to identification, and if achieved successfully can dramatically reduce the amount of required labeling effort.
Clustering is also an interesting problem in itself as it can answer questions such as who are the main characters, or what are their social interaction groups.

\begin{figure}[t]
\centering
\includegraphics[width=0.9\linewidth]{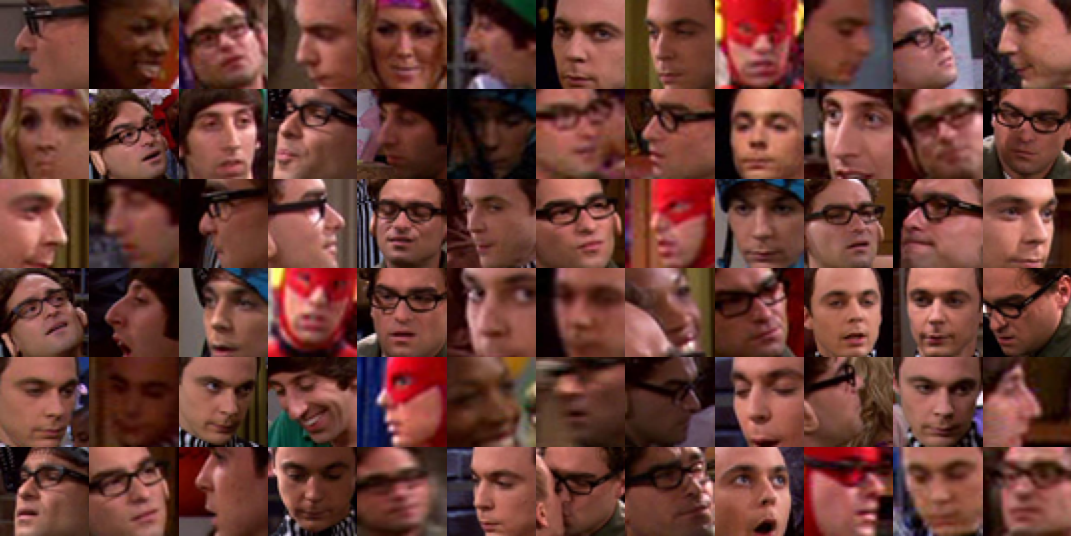}
\caption{Video face clustering is a challenging problem that is further accentuated by a large portion of characters that play small roles.
Can you guess how many characters are in this montage and which faces belong to them?
See Fig.~\ref{fig:bcl} for our solution.}
\label{fig:firstpage}
\vspace{-4mm}
\end{figure}

While there exists a large body of work in video face clustering (\eg~\cite{cinbis2011unsupervised,jin2017erdosclustering,zhang2016jfac}), most of it addresses a simplified setup where background characters%
\footnote{We consider three types of characters based on their roles:
\emph{primary} or \emph{recurring} characters have major roles in several episodes;
\emph{secondary} or \emph{minor} characters are named and play an important role in some episodes; and
\emph{background} or \emph{unknown} (Unk) characters are unnamed and uncredited.
}
are ignored and the total number of characters is known.
With recent advances in face representations~\cite{qiong2018vggface2}, their application towards clustering~\cite{sharma2017clustering}, and the ability to learn cast-specific metrics by looking at overlapping face tracks~\cite{cinbis2011unsupervised}, we encourage the community to address the challenging problem of estimating the number of characters and not ignoring background cast (see Fig.~\ref{fig:firstpage}).

In this paper, we propose \emph{Ball Cluster Learning} (BCL) - a supervised approach to carve the embedding space into equal-sized balls such that samples within a ball belong to one cluster.
In particular, we formulate learning constraints that create such a space and show how the ball radius (also learned) can be associated with the stopping criterion for agglomerative clustering to estimate both the number of clusters and assignment (Sec.~\ref{sec:method}).
We demonstrate BCL on video face clustering in a setup where we are unaware of the number of characters, and \emph{all} face tracks, main character or otherwise, are included (Sec.~\ref{sec:facetrack_cluster}).
Thus, BCL is truly applicable to all videos as it does not place assumptions on availability of cast lists (to determine number of clusters) or track labels (to discard background characters).
To evaluate our approach, we augment standard datasets used in video face clustering by resolving labels between all background characters.
Our approach achieves promising results in estimating the number of clusters and the cluster assignment.
We also present a thorough analysis of commonly used loss functions in verification (\eg~contrastive loss), compare them against BCL, and discuss how and when they may be suitable for clustering.
To the best of our knowledge, BCL is the first approach that learns a threshold to estimate the number of clusters at test time.
Code and data are available at Github.


\section{Related Work}
\label{sec:relwork}
\vspace{-1mm}
We survey work on identifying and clustering characters in videos.
We also review metric learning approaches, some of which are adopted for clustering in this work (Sec.~\ref{subsec:baselines}).

\vspace{-5mm}
\paragraph{Character identification in videos.}
Over a decade earlier, the availability of transcripts (speaker names and dialogs) and their alignment with subtitles (dialogs and timestamps) opened exciting avenues for fully automatic identification~\cite{baeuml2013semisupervised,everingham2006,ramanathan2014coref,sivic2009}.
Dialog-based supervision
proved to be a harder but scalable approach~\cite{cour2010subtitle,haurilet2016naming}.
Face track representations (\eg~\cite{liu2017qan,parkhi2014facetrackfv,parkhi2015vggface,wang2012cdl,yang2017nan}) further improved performance.
Recently, the source of supervision moved towards web images from IMDb~\cite{aljundi2016imdb,vicol2018moviegraphs} or image search~\cite{nagrani2017sherlock}, and a combination of modalities such as hair~\cite{nagrani2017sherlock}, speech~\cite{nagrani2018pin} and clothing~\cite{tapaswi2012bbt}.
However, these advances are limited to identifying named characters and grouping all remaining characters in a common ``\emph{others}" label.

\vspace{-5mm}
\paragraph{Video face clustering.}
A common idea adopted by many clustering approaches is to use unsupervised constraints that arise from the video to learn cast-specific metrics~\cite{cinbis2011unsupervised}.
Pairs of face images within tracks are considered similar; and faces that appear simultaneously in the video are assumed dissimilar.
These constraints are used with Hidden Markov Random Fields~\cite{wu2013simultaneous,wu2013constrained},
or to learn low-rank block-sparse representations~\cite{xiao2014weighted}.
They also see use in conjunction with the video editing structure (shots, threads, and scenes)~\cite{tapaswi2014facecluster}.
The constraints are also used to fine-tune CNNs and learn clustering jointly~\cite{zhang2016jfac},
or to learn an embedding using an improved triplet loss~\cite{zhang2016imptriplet}.

Ignoring tracks, metrics are learned by ranking a batch of frames and creating hard positive and negative pairs~\cite{sharma2019ssiam}.
However, all of the above methods require knowledge of the number of clusters $K$ that is difficult to estimate beforehand; and only consider primary characters (tracks for background characters are ignored).
In online face clustering spatio-temporal constraints along with CNN representations are used to assign a new track to existing or new cluster~\cite{kulshreshtha2018online}.
However, only primary characters in the video are targeted.
Recently, an end-to-end detection and clustering approach considers false positive and missed detections~\cite{jin2017erdosclustering}.

In this paper, we consider a setup where \emph{all} face tracks are to be clustered into an \emph{unknown} number of characters.

\begin{figure}[t]
\begin{center}
\includegraphics[width=0.76\linewidth]{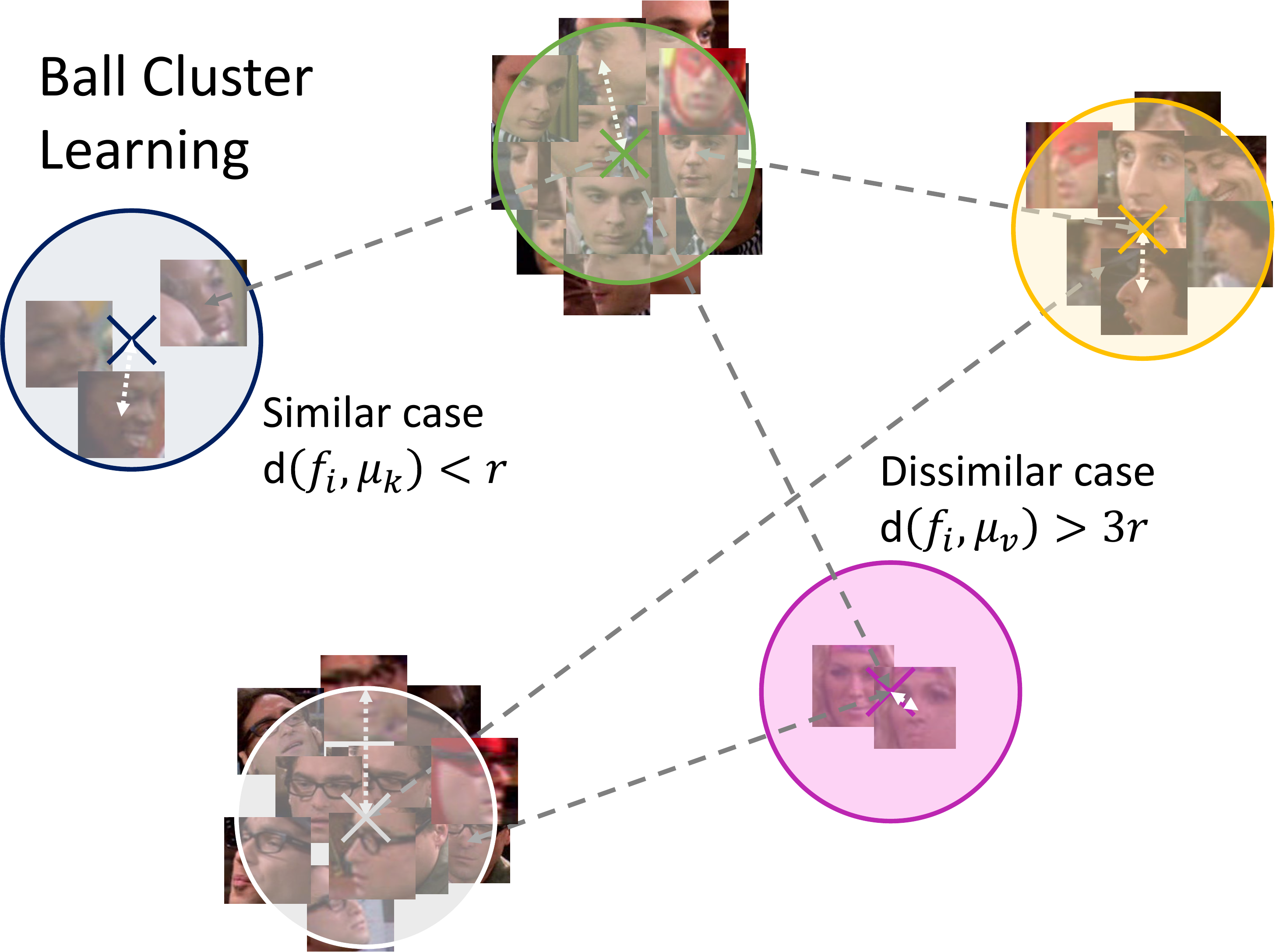}
\end{center}
\vspace{-3mm}
\caption{The face tracks in Fig.~\ref{fig:firstpage} can be clustered into $5$ characters.
\emph{Ball Cluster Learning} carves the feature space into balls of equal radius.
The number of samples in the cluster does not affect the ball radius or minimum separation to other balls.}
\label{fig:bcl}
\vspace{-6mm}
\end{figure}

\vspace{-5mm}
\paragraph{Metric learning.}
Early examples of state-of-the-art approaches in face recognition adopt metric learning~\cite{chopra2005learning,guillaumin2009ldml}. 
The learning task is often posed as verification - are two face images of the same person.
The main difficulty is ensuring that the model generalizes to test images of people that are not seen during training.
When working with videos, this is mitigated by obtaining positive and negative pairs through tracking and training on the video itself.

Other loss functions involving \textit{triplets}~\cite{schultz2004learning}  
are also proposed for face verification~\cite{schroff2015facenet}.
While FaceNet~\cite{schroff2015facenet} claims to be good at clustering, performance is only evaluated qualitatively.
Training with the triplet loss is cumbersome as it requires creation of all possible triplets that is computationally expensive.
Sampling strategies become crucial to ensure fast convergence while avoiding degenerate solutions.

Centroid-based losses~\cite{law2018dimensionality,pmlr-v70-law17a,mensink2012metric,snell2017prototypical} are also proposed for face verification~\cite{wang2017normface}.
Here, models are trained to make each sample closer to the representative of its category than to the representative of any other category resulting in samples of the same category being grouped into a single cluster.
These methods are ideal when the number of clusters is known at test time.
However, there is neither a constraint on the size/radius of the clusters,
nor is there a threshold to predict whether two samples are similar,
\eg~NormFace~\cite{wang2017normface} trains a classifier to determine whether pairs are similar.

Joint Unsupervised LEarning (JULE)~\cite{yang2016jule} learns representations while performing hierarchical clustering.
However, as JULE has to learn both the cluster assignments and representations, it is hard to scale~\cite{ghasedi2017deep,guerin2017cnn,min2018survey} and its computational cost and memory complexity are extremely high.
Moreover, JULE is only tested in cases where the number of clusters is known at test time.

We propose a model that groups similar samples into non-overlapping balls.
The radius of the ball clusters is learned and is directly related to the threshold used as a stopping criterion of our clustering algorithm (Sec. \ref{subsec:hac}).
In addition, our training algorithm has very low algorithmic complexity: it is linear in the batch size and in the number of clusters.


\section{Ball Cluster Learning (BCL)}
\label{sec:method}

The main goal of our supervised learning approach is to carve the embedding space into balls with a shared but trainable radius for each cluster, while simultaneously creating a well-defined separation between balls of different cluster labels (Fig.~\ref{fig:bcl}).
We first define the constraints that achieve the above goal (Sec.~\ref{subsec:constraints}), formulate the learning problem with loss functions (Sec.~\ref{subsec:losses}), and then explain how to perform clustering at test time (Sec.~\ref{subsec:hac}).
Finally, we review several losses from the metric learning literature that may be suitable for clustering (Sec.~\ref{subsec:baselines}).

\vspace{-5mm}
\paragraph{Notation.}
Let $B = \{ (x_i, y_i) \}_{i=1}^N, y_i \in \{ 1, \cdots, K \}$ be a mini-batch containing $N$ samples that we wish to group into $K$ clusters.
We learn a mapping $\varphi_{\theta} : \Xcal \to \Fcal$ (\eg a neural network) parameterized by $\theta$.
The embedding space can either be $\Fcal = \RR^D$ or $\Fcal = \SSD = \{ \fvec \in \RR^D : \| \fvec \|_2 = 1 \}$ inspired by recent work~\cite{wang2017normface} that shows benefits of $\ell_2$ normalization for face recognition.
The $i$-th sample is represented by the output of the mapping $\fvec_i = \varphi_{\theta} (x_i)$.

\textbf{``Ball'' terminology}: We define samples of our clusters as lying in a ball.
However, when $\Fcal = \SSD$, our clusters technically lie on the hypersurface of \emph{hyperspherical cones}.

\subsection{Constraints}
\label{subsec:constraints}

We analyze similar and dissimilar samples separately.
Let $\Ccal_k$ be the $k$-th set of similar samples (\ie samples $x_i$ that satisfy $y_i = k$).
A pair of samples $(x_i,x_j)$ is similar iff $y_i = y_j$, and it is dissimiliar otherwise.

\vspace{-5mm}
\paragraph{Similar case.}
We define $\muu_k \in \Fcal$ as the centroid of all the samples in $\Ccal_k$ \wrt the squared Euclidean distance:
\vspace{-1mm}
\begin{equation}
\muu_k = \frac{ 1 }{\ \nu_{\Fcal}} \sum_{x_i \in \Ccal_k} \fvec_i \in \argmin_{\muu \in \Fcal} \sum_{x_i \in \Ccal_k} \dsf^2(\fvec_i, \muu),
\vspace{-1mm}
\end{equation}
where $\dsf : \Fcal \times \Fcal \to \RR$ is the Euclidean distance (\ie $\dsf^2(\fvec_i, \muu_k) = \| \fvec_i - \muu_k \|_2^2$).
The factor $\nu_{\Fcal}$ is $N$ if $\Fcal = \RR^D$, or ${\|  \sum_{x_i \in \Ccal_k} \fvec_i \|_2}$ (we assume for simplicity that it is non-zero) if $\Fcal = \SSD$ since $\muu_k$ is constrained to be in $\Fcal$ (on the unit-norm hypersphere).
For any sample $x_i$ that belongs to $\Ccal_k$, we would like to learn a representation $\fvec_i$ such that its squared distance to $\muu_k$ is smaller than some learned threshold $b > 0$.
Our goal is to satisfy the constraints:
\vspace{-1mm}
\begin{equation}
\label{eq:similar_constraint}
\forall x_i \in \Ccal_k, ~ \dsf^2(\fvec_i, \muu_k) \leq b .
\vspace{-1mm}
\end{equation}
Note that $b$ is trained as a model parameter.
We consider that the \textit{radius} $r$ of the balls is
$r = \sqrt{b} \geq \max_{x_i \in \Ccal_k} \dsf(\fvec_i, \muu_k)$.
By using the triangle inequality, similar samples satisfy the following constraint:
\vspace{-1mm}
\begin{equation}
\label{eq:similar_pair_constraint}
\forall x_i \in \Ccal_k ,x_j \in \Ccal_k, \,\, \dsf(\fvec_i, \fvec_j) \leq 2r = 2 \sqrt{b} .
\vspace{-1mm}
\end{equation}
We choose $2 r$ as the threshold to determine whether two samples are similar or not. 



\begin{figure}[t]
\vspace{-2mm}
\includegraphics[trim={1.0cm 1.6cm 1.5cm 1.6cm},clip,height=2.6cm]{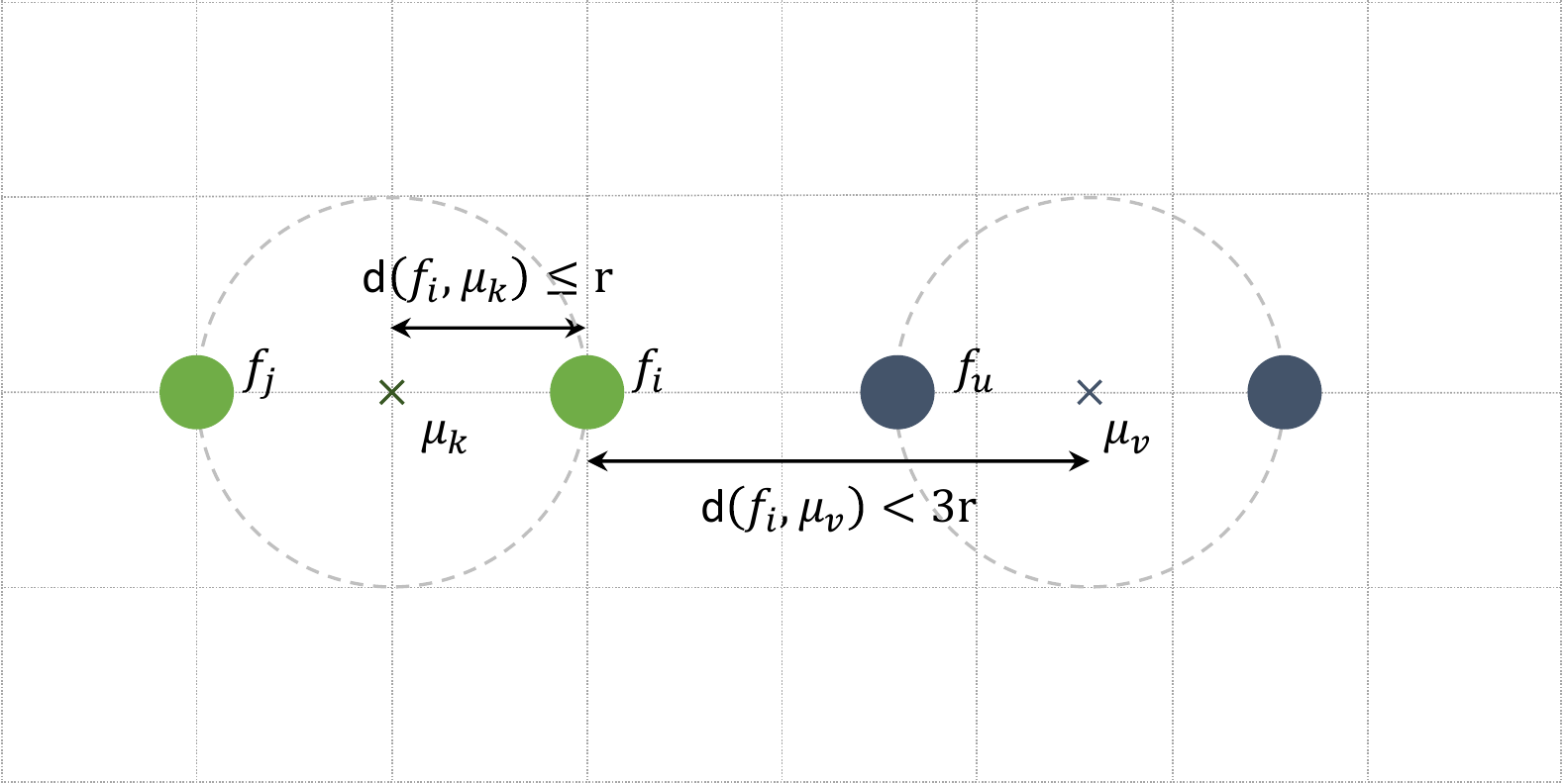} \\
\includegraphics[trim={1.0cm 1.6cm 0.5cm 1.6cm},clip,height=2.6cm]{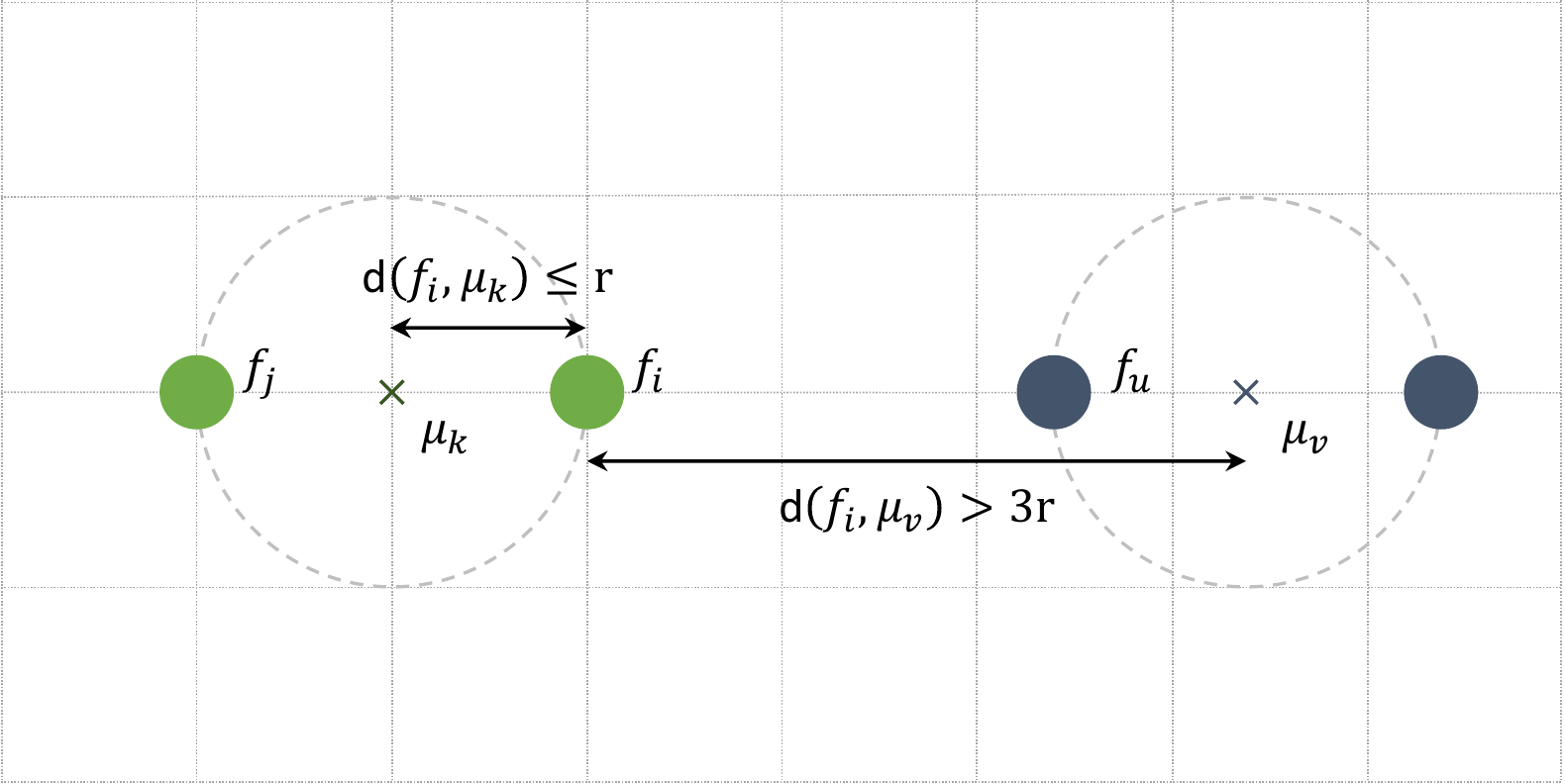}
\vspace{-2mm}
\caption{
Consider a toy scenario with 4 samples (2 green, 2 blue in each cluster) in $\Fcal = \RR^2$.
We illustrate the constraints derived in Eq.~\eqref{eq:similar_constraint} and Eq.~\eqref{eq:dissimilar_constraint}.
Each grid square in this 2D-space corresponds to the ball radius $r$.
\textbf{Top}: When $\dsf(f_i, \mu_v) < 3r$, we see that $\fvec_i$ and $\fvec_u$ are the closest samples and will be merged by hierarchical agglomerative clustering (HAC) in the first iteration.
\textbf{Bottom}: When $\dsf(f_i, \mu_v) > 3r$, the distance between $\fvec_i$ and $\fvec_u$ is larger than either the green or blue pair of samples.
Additionally, by adopting the max linkage and choosing the stopping criterion for HAC as $\tau = 2r$ (in Euclidean distance), iterative merging stops after the green and blue samples are grouped.
Best seen in color.
}
\vspace{-3mm}
\label{fig:model}
\end{figure}

\vspace{-5mm}
\paragraph{Dissimilar case.}
From the above discussion, two dissimilar samples $(x_i,x_u)$ should satisfy $\dsf(\fvec_i, \fvec_u) > 2 r$.
Furthermore, as the distance between $x_u \in \Ccal_v$ and its centroid $\muu_v$ is at most $r$, the Euclidean distance between $\fvec_i$ and $\muu_v$ should be greater than $3r$ to ensure that all the clusters are separated (see Fig.~\ref{fig:model}).
This implies $\dsf^2(\fvec_i, \muu_v) > (3r)^2 = 9b$.
We denote $\gamma = 9b + \varepsilon$ where $\varepsilon \geq 0$ is a small fixed margin and formulate the constraint:
\begin{equation}
\label{eq:dissimilar_constraint}
\forall x_i \in \Ccal_k \neq \Ccal_v, \,\, \dsf^2(\fvec_i, \muu_v) \geq \gamma .
\end{equation}

A major difference to existing metric learning approaches is that we enforce an upper bound on the distance between each example and its desired centroid (Eq.~\eqref{eq:similar_constraint}), which in turn enforces samples of each cluster to be within a ball of radius $r$.
We also enforce different clusters to be separated by a margin that is a function of the radius (Eq.~\eqref{eq:dissimilar_constraint}).

\vspace{-5mm}
\paragraph{Computational complexity.}
Formulating our constraints on the distances between samples and cluster centroids significantly lowers the number of computations in contrast to pairwise distances that yield quadratic constraints.

\vspace{-5mm}
\paragraph{A fixed radius}
for all balls allows us to use it as a threshold to delimit clusters.
In addition, it has the potential to address the long-tail since each identity gets the same volume of embedding-space, agnostic to the number of tracks.


\subsection{Problem Formulation}
\label{subsec:losses}

Based on the desired constraints in Sec.~\ref{subsec:constraints}, we now formulate an optimization problem that tries to satisfy them.
Our goal is to learn the squared radius $b > 0$ of the cluster balls and parameters $\theta$ of the model $\varphi_\theta$ that minimize the objective problem $\mcL_{ball}$ defined as the sum of the two losses:
\begin{equation} \label{eq:bcl_problem}
\mcL_{ball} = \alpha \mcL_{sim}  + \mcL_{dis} ,
\end{equation}
where $\alpha \geq 0$ is a hyperparameter to balance the losses.
We present details of the loss terms in the following.



The goal of the loss $\mcL_{sim}$ is to satisfy the similar pairs constraint in Eq.~\eqref{eq:similar_constraint}, and is formulated as:
\begin{equation}
\mcL_{sim} = \frac{1}{N} \sum_{x_i \in \Ccal_k}  \left[ \dsf^2( \fvec_i , \muu_k)  - b \right]_+ ,
\end{equation}
where $\left[ x \right]_+ = \max(0, x)$.
In the context of metric learning, this often corresponds to the \emph{positive} loss as it brings together samples of the same cluster.

The goal of the dissimilar loss $\mcL_{dis}$ is to satisfy dissimilar pairs constraints in Eq.~\eqref{eq:dissimilar_constraint} and is formulated as:
\begin{equation}
\mcL_{dis} = \frac{1}{N} \sum_{x_i \in \Ccal_k} \max_{v \neq k} \left[ \gamma - \dsf^2 ( \fvec_i , \muu_v )  \right]_+ .
\end{equation}
This loss aims to push away from the \emph{most} offending cluster centroid by employing $\max_{v \neq k}$, and is equivalent to hard negative mining in metric learning~\cite{schroff2015facenet}.

\subsection{Clustering Algorithm}
\label{subsec:hac}

We now describe how to perform clustering and predict the number of clusters on some given (test) dataset.
Recall that we are interested in solving problems where the number of clusters is unknown at test time.

As explained in Sec.~\ref{subsec:constraints}, our constraints are formulated so that similar samples should satisfy $\dsf^2(\fvec_i, \fvec_j) \leq 4 b$ and dissimilar samples should have larger distances.
We apply a clustering algorithm which groups pairs of examples that satisfy those constraints into a single cluster.

Even when the number of clusters is known, finding the partitions that minimize some clustering energy function is an NP-hard problem~\cite{aloise2009np}.
Thus, methods that find a good local minimum solution with reasonable complexity are often used (\eg $K$-means~\cite{lloyd1982least}).
For this reason, we adopt the Hierarchical Agglomerative Clustering (HAC) method~\cite{defays1977efficient}: each sample starts in its own cluster, and pairs of clusters are iteratively merged until some specific stopping criterion.
In the context of \textit{complete linkage}, two clusters $U$ and $V$ are merged into a single cluster if they minimize:
\begin{equation} \label{eq:hac_aggregation}
  \ell_{complete} (U, V) =  \max_{x_u \in U, x_v \in V} ~ \dsf^2(\fvec_u, \fvec_v) .
\end{equation}
Let us denote $\tau > 0$ the threshold chosen such that the HAC algorithm stops when there are \emph{no} two clusters $U$ and $V$ that satisfy $\ell_{complete}(U, V) \leq \tau$.
Once the HAC algorithm stops, all the examples assigned to a same cluster $U$ satisfy
\mbox{$\forall x_a \in U,$} $x_b \in U, ~ \dsf^2(\fvec_a, \fvec_b) \leq \tau$ by definition of the complete linkage.
Thus, we choose $\tau = 4b$. 
With this value of $\tau$, when the (ideal) global minimum of Eq.~\eqref{eq:bcl_problem} is obtained, applying the HAC with linkage in Eq.~\eqref{eq:hac_aggregation} groups similar examples in the same clusters and separates dissimilar examples since both Eq.~\eqref{eq:similar_pair_constraint} and  Eq.~\eqref{eq:dissimilar_constraint} are satisfied.

\subsection{Extending related work to our task}
\label{subsec:baselines}

We compare BCL with various metric learning approaches commonly used in face verification tasks.

\vspace{-4mm}
\paragraph{Triplet Loss~\cite{schroff2015facenet}} tries to preserve the order of distances between similar pairs $(x_i, x_j)$ and dissimilar pairs $(x_i, x_u)$:
$\mcL_{triplet} = \sum_{\substack{y_i = y_j \\ y_i \neq y_u}} \left[ \dsf^2(\fvec_i, \fvec_j) - \dsf^2(\fvec_i, \fvec_u) + m \right]_+ $.
While the loss ensures that positives are closer than negatives by a margin $m$, there is no constraint on the distance between positive samples.
Thus, we are unable to directly use the margin as a threshold for stopping the HAC algorithm.

\vspace{-4mm}
\paragraph{Threshold strategy.}
We choose a threshold based on a validation set: we apply the HAC algorithm and pick the threshold that predicts the ground truth number of validation clusters.
Even for the baselines that learn a threshold, this strategy worked better than using the learned threshold.
Thus, we report scores using this strategy for all baselines.

\vspace{-4mm}
\paragraph{Contrastive Loss~\cite{chopra2005learning}} considers pairwise constraints.
For any pair of samples $(x_i, x_j)$, and $y_{ij} = 1$ when they are similar and $0$ otherwise, the contrastive loss between them is
$\mcL_{cont} =  \frac{y_{ij}}{2}  \dsf^2(\fvec_i, \fvec_j) + \frac{1}{2} (1 - y_{ij})  \left[ m - \dsf(\fvec_i, \fvec_j) \right]^2_+$.
This aims to make dissimilar samples at least $m$ distance apart.
While $m$ is usually a fixed hyperparameter, we treat it as a trainable value in the same way as $b$ in BCL.

\vspace{-4mm}
\paragraph{Logistic Discriminant Metric Learning (LDML)~\cite{guillaumin2009ldml}}
maps distances to a probability score via the sigmoid function $\sigma(\cdot)$.
It can be written as
$p_{ij} = p(y_i = y_j|\fvec_i, \fvec_j, \beta) = \sigma \left(\beta - \dsf^2(\fvec_i, \fvec_j) \right)$,
where $\beta$ is a threshold trained to distinguish similar from dissimilar pairs.
The loss is formulated as binary cross-entropy and is minimized:
$\mcL_{ldml} = - \sum_{y_i = y_j} \log p_{ij} - \sum_{y_i \neq y_j} \log (1 - p_{ij})$.


\vspace{-4mm}
\paragraph{Prototypical Networks~\cite{snell2017prototypical}}
If both Eq. \eqref{eq:similar_constraint} and Eq. \eqref{eq:dissimilar_constraint} are satisfied, the following order is obtained:
$\forall x_i \in \Ccal_k$, \mbox{$v \neq k$,} $\dsf^2(\fvec_i, \muu_k) - b \leq \dsf^2(\fvec_i, \muu_v) - \gamma$.
To satisfy this relative constraint, we formulate the cross entropy loss:
\begin{align} \label{eq:proto}
\mcL_{proto} = - \frac{1}{N} \sum_{i \in \Ccal_k} \frac{1}{| \Ccal_k |} \log \left( p(y_i = k | \fvec_i ) \right)
\end{align}
where $p(y_i = k | \fvec_i )$ is the posterior probability:
\begin{equation}
\frac{\exp(-\dsf^2 (\fvec_i , \muu_k) + b)}{\exp(-\dsf^2 (\fvec_i , \muu_k) + b) + \sum_{v \neq k} \exp(-\dsf^2 (\fvec_i , \muu_v) + \gamma)} .
\end{equation}
The vanilla Prototypical Networks~\cite{snell2017prototypical} correspond to $\mcL_{proto}$ when $b = \gamma = 0$.
NormFace~\cite{wang2017normface} is similar to~\cite{snell2017prototypical}, with one main difference that representations are $\ell_2$-normalized.
We report scores for $b = \gamma = 0$ since we experimentally found that it returns the best results with our threshold strategy.


\section{Video Face Track Clustering with BCL}
\label{sec:facetrack_cluster}

We discuss how BCL can be applied to face track clustering.
Each sample represents a face track and is associated with a specific identity.
Our goal is to create clusters such that tracks with the same identity are grouped together.

\begin{figure}[t]
\begin{center}
\includegraphics[width=\linewidth]{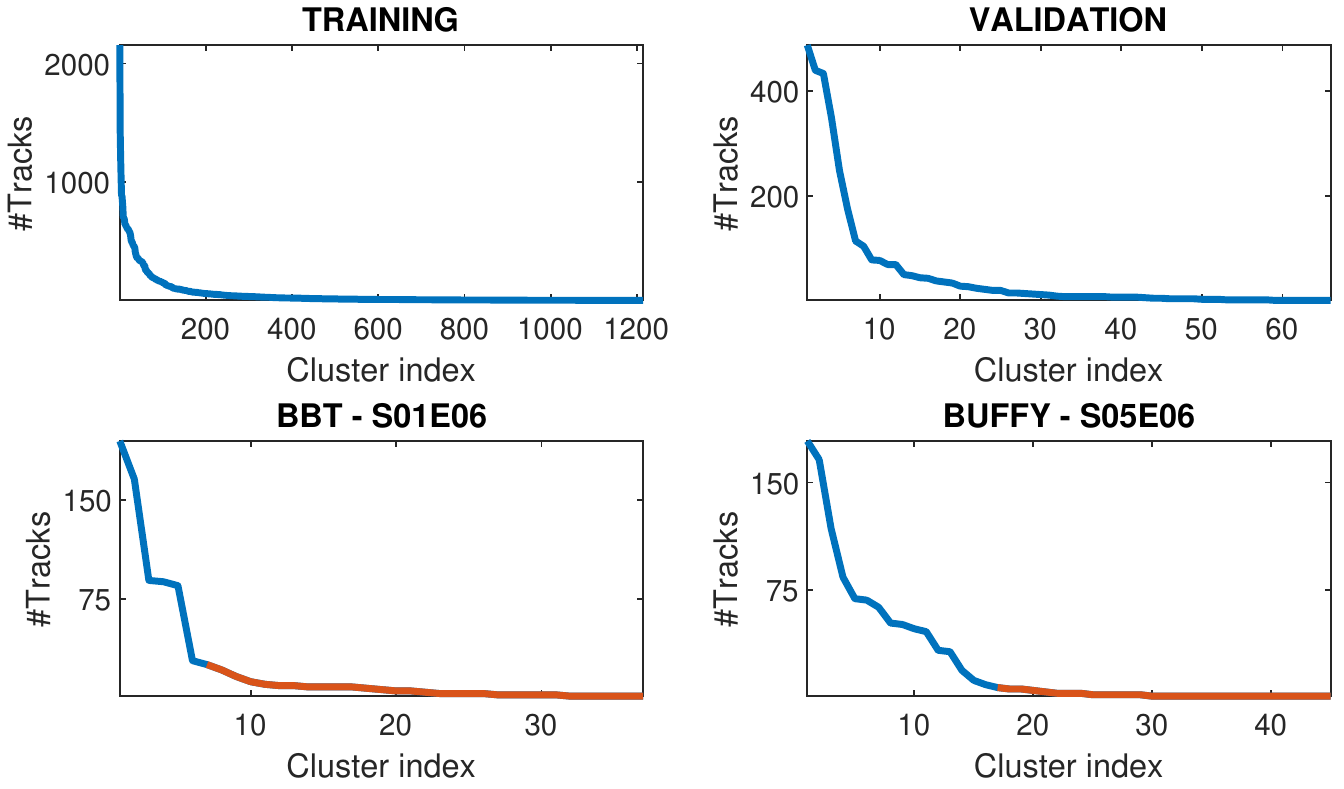}
\end{center}
\vspace{-3mm}
\caption{Number of tracks in clusters.
Orange lines in BBT and BUFFY indicate track counts for unknown/background characters.}
\vspace{-5mm}
\label{fig:skewed_clusters}
\end{figure}

During training, we create mini-batches by uniformly sampling a fixed number of tracks.
As the training data contains several identities with very few (1-2) tracks, and many others with hundreds or thousands of tracks, uniform random sampling preserves the skewed distribution of cluster membership within the mini-batch (see~Fig.~\ref{fig:skewed_clusters}).
From each track, we randomly choose one face image (which serves as data augmentation) and use a pre-trained and fixed CNN to extract a face representation $x_i$.
We will refer to this as the base CNN representation.
At test time, we average the base representations of all face images in the track and apply HAC after computing embeddings.
This makes the track feature robust, while keeping it in the same space as the training samples.
Other track-level representations such as~\cite{liu2017qan,parkhi2014facetrackfv,wang2012cdl,yang2017nan} are out of scope of this work.

\vspace{-5mm}
\paragraph{Base CNN}
is a 50-layer ResNet~\cite{he2016resnet} with squeeze-and-excitation (SE) blocks~\cite{hu2018senet}.
The model is pre-trained on the MS-1M dataset~\cite{guo2016msceleb}, and fine-tuned on the VGGFace2 dataset~\cite{qiong2018vggface2} with cross-entropy loss to predict over 8000 identities.
We obtain features in $\mathbb{R}^{256}$ from the last layer (before the classifier).
This model is named SE-ResNet50-256.\footnote{We use the pre-trained PyTorch model provided by \url{https://github.com/ox-vgg/vgg_face2}.}
We will show that our methods work equally well when using a different base CNN.
We do not fine-tune the CNN.

\vspace{-5mm}
\paragraph{Model.}
Our model $\varphi_\theta$ is a stack of 4 linear layers with ReLU non-linearity (MLP) in between and is applied on top of the base CNN representation.
When not stated otherwise, the hidden layers have 256, 128, and 64 nodes, and the final embedding dimension $D = 64$.

Our constraints require that $b > 0$.
To this end, we use the \texttt{softplus} operator defined as $b = \log(1 + e^{\hat{b}})$, and train $\hat{b} \in \RR$ as a model parameter.
We balance the similar and dissimilar losses with $\alpha=4$ based on the performance on a validation set.

\vspace{-5mm}
\paragraph{Learning.}
We find that the loss for our model can be reduced dramatically (to $\epsilon$) by mapping all samples to the same point and learning the squared radius to be close to 0.
We prevent the learning process from reducing the radius to 0, by freezing it for the first 5 epochs.
Subsequently, the loss parameter $\hat{b}$ updates slowly, 0.1 times the learning rate used for MLP weights.
We employ SGD with 0.9 momentum at a learning rate of 0.003, and a 0.9$\times$ decay every 10 epochs to update the weights of the MLP.
We use mini-batches of 2000 samples (tracks) when not stated otherwise.


\section{Evaluation}
\label{sec:eval}

We first present the datasets and metrics used in our experiments.
Then, we perform ablation studies on the validation split and finally show and discuss our results on the test set.

\subsection{Datasets and metrics}
\label{subsec:eval:datasets}
We use face tracks from several movies and TV series as part of training and evaluation.

\vspace{-5mm}
\paragraph{Train and validation splits}
consist of face tracks and ground-truth identity labels provided for 51 movies from the MovieGraphs dataset~\cite{vicol2018moviegraphs}.
Like most previous work, the dataset contains annotations for main characters only, and does not disambiguate between background characters.
Nevertheless, it is suitable for training, and we obtain 65,076 tracks that are mapped to 1,280 unique actors using IMDb.
As Fig.~\ref{fig:skewed_clusters} indicates, many actors have few (and even one) tracks making the training distribution similar to test.

We reserve 5\% of the actors for validation and ensure that actors appearing in the test data are \emph{not} seen during train or validation.
This results in 61,774 tracks (1,214 actors) for the train split and 3,302 tracks (66 actors) for validation.

\vspace{-5mm}
\paragraph{Test split.}
Our evaluation is on six episodes each of two TV series: The Big Bang Theory (BBT) and Buffy the Vampire Slayer (BUFFY).
Both have been actively used in person identification and clustering~\cite{baeuml2013semisupervised,jin2017erdosclustering,sharma2019ssiam,zhang2016jfac}.\footnote{We use an updated version of the tracks that does not discard background characters and small/profile faces.}
We wish to emphasize that most previous approaches for face clustering only consider primary (recurring) characters and know the number of clusters.
We adopt a more practical setting where the number of characters is \emph{not} known, and tracks for \emph{all} (secondary as well as background characters) are included.
We painstakingly resolve faces of background characters and assign unique identifiers to them.
This is difficult even for humans, but is achieved through a combination of facial (hair) and non-facial (clothing, spatial location) cues.
Finally, we also evaluate on \emph{combined tracks} from several episodes (and series) to mimic additional challenging scenarios.
Each face in Fig.~\ref{fig:firstpage} represents a different track from BBT-S1E1.

\vspace{-5mm}
\paragraph{Metrics.}

We adopt three primary metrics to evaluate performance:
(i) $\#$Cl: is the number of predicted clusters, and should be close to the ground-truth number of identities. \\
(ii) Normalized Mutual Information (NMI)~\cite{manning_irbook}: for a given set of class labels $Y$ and cluster predictions $C$, NMI is calculated as $2 I(Y; C) / (H(Y) + H(C))$, where $H(\cdot)$ is entropy, and $I(\cdot; \cdot)$ is mutual information.
NMI is a balanced metric and scores 0 when all samples are either in one cluster or their own clusters (see Table~\ref{table:validation_ball_clustering}).
All model checkpoints are chosen to maximize NMI on the validation set. \\
(iii) Weighted Clustering Purity (WCP)~\cite{tapaswi2014facecluster}: Also called clustering accuracy~\cite{zhang2016jfac}, WCP combines purity (fraction of samples that belong to the same class) of the clustering by weighting with the number of samples in the cluster.


\subsection{Ablation Studies}
\label{subsec:eval:ablation}

We make several design choices that are motivated in the following.
Table~\ref{table:validation_ball_clustering} provides insights into the validation split by showing the extreme ends of the clustering.
We also demonstrate that our model outperforms the base CNN descriptors even when the base model is assumed to know the actual number of clusters (Base K known).
Throughout this section, the ideal number of clusters on validation is 66.


\begin{table}[!t]
\small
\centering
\begin{tabular}{lcccc}
\toprule
    &  One     &  N        &  Base           &    Ours      \\
    &  cluster & clusters  & $K$ known       & $\tau = 4 b$ \\
\midrule
$\#$Cl        &   1    &  3302  &    66   &    69   \\
NMI           &   0    &    0   &  68.91  &  77.09  \\
WCP           & 14.75  & 100.0  &  76.53  &  85.65  \\
\bottomrule
\end{tabular}
\caption{Performance on the validation set showing the impact of putting all samples in the same or their own clusters.
We also present performance of base CNN features when the number of clusters is known.
The validation set has 66 ground-truth clusters.}
\vspace{-2mm}
\label{table:validation_ball_clustering}
\end{table}

\begin{table}[!t]
\small
\centering
\tabcolsep=0.16cm
\begin{tabular}{lccccc}
\toprule
Base CNN           & Dim   & $\#$P & $\#$Cl &   NMI   &   WCP   \\
\midrule
SE-ResNet50-256    & 256   & 26.5M &   69   &  \textbf{77.09}  &  85.65  \\
ResNet50           & 2048  & 41.1M &   80   &  76.74  &  \textbf{87.67}  \\
\bottomrule
\end{tabular}
\caption{Performance on validation set for different base CNN models.
The 4-layer MLP used for the ResNet50 model is 2048$\rightarrow$512$\rightarrow$256$\rightarrow$128$\rightarrow$64.}
\vspace{-2mm}
\label{table:vgg2face}
\end{table}

\begin{table}[!t]
\small
\centering
\tabcolsep=0.16cm
\begin{tabular}{lcccccc}
\toprule
Dim          &  256  &  128  &   64  &   32   &   16  &    8  \\
$\#$P        &  263K &  132K &  111K &  109K  &  108K &  107K \\
\midrule
$\#$Cl       &   45  &   62  &   69  &   \textbf{68}  &   72  &   29  \\
NMI          & 76.68 & 76.89 & \textbf{77.09} & 75.48 & 68.72 & 47.62 \\
WCP          & 81.98 & 85.52 & 85.65 & \textbf{85.89} & 79.35 & 50.79 \\
\bottomrule
\end{tabular}
\caption{Performance on the validation set for varying embedding dimension.
$\#$P indicates the number of parameters in the MLPs.}
\vspace{-4mm}
\label{table:embed_dim}
\end{table}

\begin{table}[t]
\small
\centering
\tabcolsep=0.13cm
\begin{tabular}{clcccc}
\toprule
          & Batch size     &  500  &  1000 &  2000 &  4000 \\
\midrule
$\#$Cl    & in batch       &  220  &   330 &   450 &   600 \\
(approx)  & $>5$ samples   &   15  &    45 &    90 &   150 \\
\midrule

Performance & $\#$Cl       &   88  &   91  &  \textbf{69}   &   29  \\
  on        & NMI          & 72.13 & 74.63 & \textbf{77.09} & 76.55 \\
Validation  & WCP          & 83.77 & \textbf{87.28} & 85.65 & 79.68 \\
\bottomrule
\end{tabular}
\caption{Ablation studies on mini-batch size.
The first half of the table reports the number of the clusters in the batch, and those that have more than 5 samples.
In the second half, we report performance on validation.
}
\vspace{-4mm}
\label{table:batch_size}
\end{table}



\begin{table*}
\centering
\scriptsize
\tabcolsep=0.17cm
\begin{tabular}{rl|cccccc|cccccc|ccc}
\toprule
    & & \multicolumn{6}{c|}{BBT}  & \multicolumn{6}{c|}{BUFFY} & BBT & BUFFY & BOTH \\
    & &  S1E1  &  S1E2  &  S1E3  &  S1E4  &  S1E5  &  S1E6  &  S5E1  &  S5E2  &  S5E3  &  S5E4  &  S5E5  &  S5E6  & 6 ep. & 6 ep. & 12 ep.  \\
\midrule
 1 & $\#$Ch      &   8 &   6 &  26 &  28 &  25 &  37 &  13 &  22 &   15 &  32 &  38 &   45 &  103  &   109  &   212  \\
 2 & $\#$Named Ch &   6 &   5 &   7 &   8 &   6 &   6 &  11 &  12 &   13 &  14 &  13 &   17 &   11  &    26  &    37  \\
 3 & $\#$Unk Ch  &   2 &   1 &  19 &  20 &  19 &  31 &   2 &  10 &    2 &  18 &  25 &   28 &   92  &    83  &   175  \\
 4 & $\#$T       & 656 & 615 & 660 & 613 & 524 & 840 & 795 & 993 & 1194 & 898 & 840 & 1112 & 3908  &  5832  &  9740  \\
 5 & $\#$Named T  & 647 & 613 & 562 & 568 & 463 & 651 & 786 & 866 & 1185 & 852 & 733 & 1055 & 3504  &  5477  &  8981  \\
 6 & $\#$Unk T   &   9 &   2 &  98 &  45 &  61 & 189 &   9 & 127 &    9 &  46 & 107 &   57 &  404  &   355  &   759  \\
\hline
\multicolumn{17}{>{\columncolor{mygray}}c}{CrossEntropy Loss} \\[0.1mm]
 7 & $\#$Cl      &  23   &  24   &  37   &  38   &  \textbf{26}   &  \textbf{37}   &  43   &  39   &  58   &  56   &  49   &  \textbf{52}   &  130  &  194  &  323  \\
 8 & NMI         & 67.42 & 64.57 & 64.87 & 69.73 & 72.52 & 63.02 & 63.14 & 59.58 & 59.07 & 61.44 & 60.52 & 61.78 & 57.91 & 55.58 & 60.33 \\
 9 & WCP         & 96.80 & 90.57 & 86.36 & 87.93 & 86.83 & 73.81 & 86.67 & 69.99 & 78.48 & 79.73 & 78.10 & 70.68 & 86.59 & 74.57 & 76.05 \\
\hline
\multicolumn{17}{>{\columncolor{mylightorange}}c}{Logistic Discriminant Metric Learning~\cite{guillaumin2009ldml}} \\[0.1mm]
10 & $\#$Cl      &  14   &  15   &  19   &  25   &  20   &  30   &  25   &  31   &  28   &  \textbf{29}   &  31   &  30   &  62   &  82   &  116  \\
11 & NMI         & 66.42 & 53.21 & 66.59 & 65.33 & 73.06 & 55.77 & 63.57 & 53.38 & 58.54 & 59.52 & 52.68 & 56.50 & 53.15 & 50.65 & 51.97 \\
12 & WCP         & 92.23 & 82.28 & 74.70 & 79.61 & 86.07 & 62.86 & 83.02 & 58.71 & 71.69 & 67.59 & 59.17 & 59.80 & 74.33 & 61.01 & 58.14 \\
\hline
\multicolumn{17}{>{\columncolor{mylightorange}}c}{Contrastive Loss~\cite{chopra2005learning}} \\[0.1mm]
13 & $\#$Cl      &  14   &  13   &  17   &  22   &  19   &  32   &  22   &  30   &  26   &  \textbf{29}   &  29   &  27   &  60   &  71   &  110  \\
14 & NMI         & 62.45 & 63.69 & 61.77 & 65.55 & 71.38 & 55.68 & 61.00 & 53.94 & 58.15 & 53.42 & 53.59 & 52.01 & 58.94 & 49.15 & 51.53 \\
15 & WCP         & 90.70 & 86.99 & 64.85 & 76.35 & 75.57 & 65.95 & 77.86 & 56.09 & 67.59 & 60.80 & 62.02 & 50.36 & 77.53 & 57.30 & 48.81 \\
\hline
\multicolumn{17}{>{\columncolor{mylightorange}}c}{Triplet Loss~\cite{schroff2015facenet}} \\[0.1mm]
16 & $\#$Cl      &  \textbf{9}    &  12   &  15   &  16   &  13   &  23   &  23   &  \textbf{24}   &  25   &  22   &  23   &  26   &  51   &  73   &  111  \\
17 & NMI         & 88.13 & 71.23 & 79.83 & 76.71 & 85.77 & 69.34 & 73.60 & 64.22 & 66.24 & 63.61 & 67.88 & 65.49 & 67.94 & 59.74 & 64.79 \\
18 & WCP         & 98.48 & 95.28 & 90.15 & 83.69 & 89.69 & 76.67 & 88.68 & 67.77 & 81.99 & 69.71 & 77.74 & 68.71 & 87.31 & 68.69 & 71.34 \\
\hline
\multicolumn{17}{>{\columncolor{mygreen}}c}{Prototypical Loss~\cite{snell2017prototypical}} \\[0.1mm]
19 & $\#$Cl      &  12   &  15   &  \textbf{22}   &  \textbf{28}   &  18   &  41   &  32   &  32   &  20   &  \textbf{35}   &  \textbf{40}   &  36   &  \textbf{87}   &  \textbf{123}  &  \textbf{197}  \\
20 & NMI         & 82.29 & 75.12 & 83.74 & \textbf{80.29} & 91.36 & \textbf{74.32} & 74.23 & 71.02 & 76.16 & 70.46 & 76.63 & 73.47 & 70.43 & 64.99 & 70.23 \\
21 & WCP         & 96.19 & 97.56 & \textbf{93.79} & \textbf{91.03} & \textbf{94.66} & \textbf{86.67} & 90.19 & \textbf{80.16} & 82.50 & 81.85 & 88.69 & 78.24 & \textbf{90.56} & 80.52 & \textbf{82.80} \\
\hline
\multicolumn{17}{>{\columncolor{mycyan}}c}{Ball Cluster Learning (Ours)} \\[0.1mm]
22 & $\#$Cl      &  \textbf{7}    &  \textbf{8}    &  16   &  18   &  11   &  23   &  \textbf{17}   &  16   &  \textbf{18}   &  22   &  26   &  22   &  47   &  71   &  116  \\
23 & NMI         & \textbf{95.81} & \textbf{87.25} & \textbf{88.38} & 76.59 & \textbf{92.21} & 74.19 & \textbf{81.78} & \textbf{77.60} & \textbf{77.64} & \textbf{78.13} & \textbf{79.72} & \textbf{78.15} & \textbf{73.22} & \textbf{71.23} & \textbf{75.32} \\
24 & WCP         & \textbf{98.63} & \textbf{98.54} & 90.61 & 86.95 & 89.12 & 81.07 & \textbf{92.08} & 79.76 & \textbf{84.00} & \textbf{84.97} & \textbf{89.05} & \textbf{80.58} & 89.36 & \textbf{83.62} & \textbf{82.81} \\
\midrule[1mm]
\multicolumn{17}{c}{Ball Cluster Learning (Ours) + Fine-tune with automatically obtained positive/negative pairs} \\
25 & $\#$Cl      &   9   &   8   &   24  &   24  &   21  &   36  &   23  &   27  &   25  &   36  &   38  &   40  &  69   &  78   &  126  \\
26 & NMI         & 97.34 & 97.80 & 94.00 & 90.42 & 95.83 & 83.32 & 84.59 & 82.59 & 78.76 & 77.58 & 81.71 & 79.51 & 88.26 & 77.05 & 80.42 \\
27 & WCP         & 99.24 & 99.67 & 96.06 & 96.08 & 97.71 & 90.36 & 94.97 & 88.12 & 90.28 & 86.19 & 90.24 & 88.13 & 94.11 & 86.64 & 85.84 \\
\bottomrule
\end{tabular}
\vspace{1mm}
\caption{Clustering performance on episodes of the test set.
S1E2 corresponds to season 1 and episode 2.
The last three columns show results on datasets created by combining tracks from several episodes.
\emph{Name} refers to primary and secondary named characters; 
\emph{Unk} refers to background characters;
$\#$Ch is number of characters;
$\#$T is number of tracks; and
$\#$Cl is number of predicted clusters and should be close to the number of characters (row 1).
Read this table by looking at each column, and seeing which method is able to predict the number of clusters and has high NMI and WCP scores.
}
\vspace{-7mm}
\label{table:test_episodes}
\end{table*}

\begin{table}
\small
\centering
\tabcolsep=0.14cm
\vspace{2mm}
\begin{tabular}{rl|ccc}
\toprule
    & & BBT & BUFFY & ALL \\
    & & 6 ep. & 6 ep. & 12 ep.  \\
\midrule
BCL  & $K$-means & 60.5 (92.0) & 66.7 (\textbf{87.3}) & 68.7 (\textbf{88.0}) \\
BCL  & HAC     & \textbf{70.6} (\textbf{93.0}) & \textbf{69.1} (85.3) & \textbf{72.5} (86.2) \\
\midrule
PRO  & $K$-means & 60.7 (91.3) & 64.5 (85.4) & 66.8 (85.6) \\
PRO  & HAC     & 68.3 (91.1) & 65.8 (80.0) & 70.3 (83.3) \\
\bottomrule
\end{tabular}
\caption{NMI and WCP performances of our approach (BCL) and prototypical loss (PRO) when the number of clusters is known.}
\vspace{-2mm}
\label{table:test_known_K}
\end{table}


\vspace{-5mm}
\paragraph{Base CNN model.}
We demonstrate that the choice of the CNN model does not directly influence performance.
In fact, our base model SE-ResNet50-256 has an output space $x_i \in \RR^{256}$ while the ResNet50 base model produces $x_i \in \RR^{2048}$.
Table~\ref{table:vgg2face} shows that both models exhibit similar performance.

\vspace{-5mm}
\paragraph{Embedding dimension.}
From the results in Table~\ref{table:embed_dim}, we can infer that choosing too small an embedding dimension reduces performance dramatically.
However, setting $D \geq 32$ achieves comparable similar performance.

\vspace{-5mm}
\paragraph{Batch size.}
Our model learns to satisfy the constraints and perform clustering on data within each mini-batch.
When batches are small (\eg~less than 50) it is likely that most clusters have only one sample.
This automatically satisfies positive constraints and gradients are 0.
Making the mini-batches too large incurs a computational cost and reduces the number of parameter updates; the model requires many more epochs to reach a similar performance.
In Table~\ref{table:batch_size}, we first report the approximate number of clusters in a batch, and the number of clusters with more than 5 samples that can be assumed to have meaningful centroids ($> 5$).
Notice how this can be quite small even for a batch of 500 samples.
We find that a batch of 2000 samples is a decent trade off that achieves good performance.

\vspace{-5mm}
\paragraph{$\ell_2$ normalized embeddings}
$\fvec_i$ (\ie $\Fcal = \SSD$) help improve performance and are used in our model.
Without the $\ell_2$ normalization, our method creates 71 clusters with NMI: 74.57 and WCP: 83.07 ($\sim$2.5\% lower).

\vspace{-5mm}
\paragraph{Single face image at training.}
We average base CNN representations of face images in a track at test time, while at training, we feed single images.
This seems conflicting.
However, when we choose to average a random half subset of track images during training, the performance is much worse with 124 clusters and a 7\% lower NMI (absolute).

\vspace{-5mm}
\paragraph{Complexity.}
During BCL training, each sample is compared only to the centers of clusters/categories.
Thus, BCL has complexity linear in the number of samples and number of categories.
This is much lower than most baselines that compare samples with samples.
We report the wall clock time (average of 3 runs) taken to compute various losses for one epoch --
Prototypical: 12.3s;
Contrastive: 15.5s;
LDML: 15.5s;
Triplet: 50.8s;
and BCL: 9.9s.

\begin{figure}[t]
\begin{center}
\includegraphics[width=0.42\linewidth]{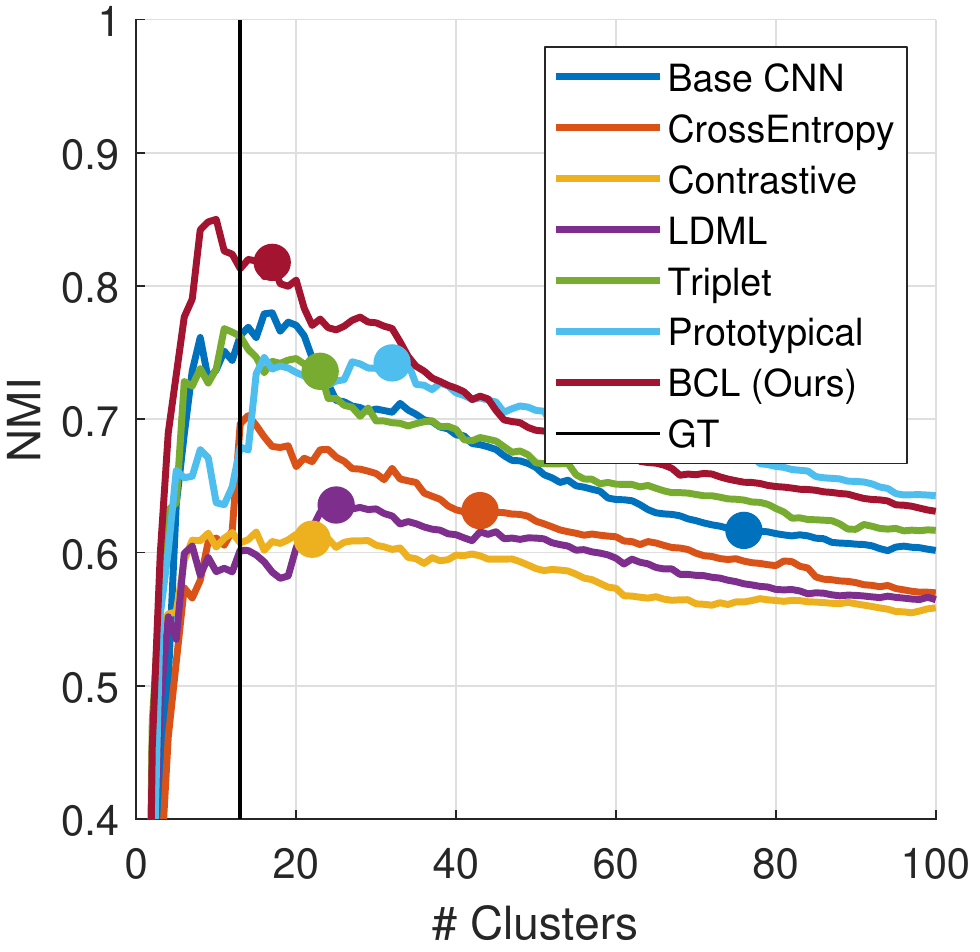} \quad
\includegraphics[width=0.42\linewidth]{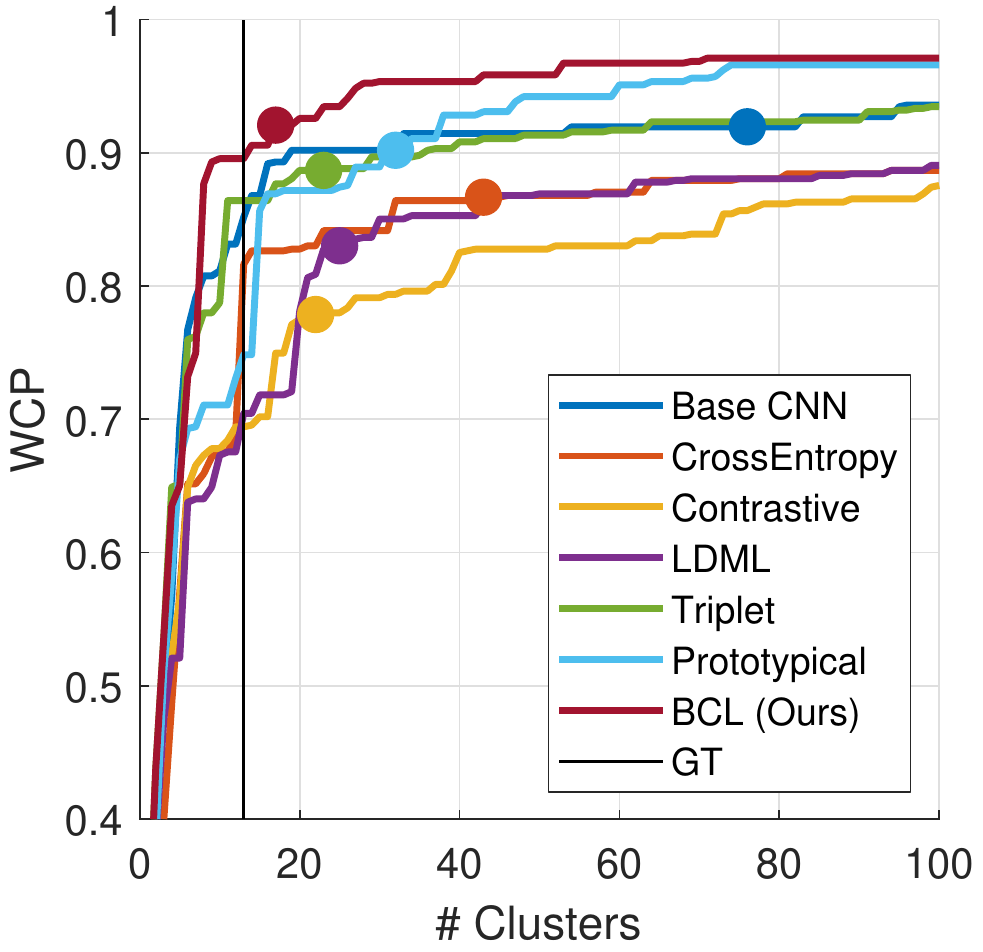}
\end{center}
\vspace{-4mm}
\caption{NMI and WCP vs. number of clusters on Buffy S5E1.
Circles indicate operating points (\ie number of predicted clusters for the methods), our method uses the HAC threshold 4$b$, while all others are using the threshold tuned to give 66 clusters on the validation set.
Best seen in color.}
\vspace{-5mm}
\label{fig:validation_losses}
\end{figure}

\subsection{Evaluation on Test Set}
\label{subsec:eval:test}

We present statistics of the test set episodes in Table~\ref{table:test_episodes}, rows 1-6.
In particular, note how some episodes have a large number of background characters (\eg~31 for BBT-S1E6) while others do not (\eg~2 for BUFFY-S5E1).
The last three columns refer to larger and arguably harder%
\footnote{The combined episode datasets have many more background characters, while tracks from recurring characters collapse onto each other.
This further skews cluster membership, with the largest cluster being several thousand tracks, and the smallest still having one track.}
datasets created by combining tracks of several episodes.
In addition to Table~\ref{table:test_episodes}, we also plot NMI and WCP vs. number of clusters in Fig.~\ref{fig:validation_losses}.
Below, we discuss each loss in detail.

\vspace{-5mm}
\paragraph{CrossEntropy loss (CE).}
CE can be seen as a (logistic) regression problem that merges all the similar examples to a single one-hot vector.
We believe this is a reason why base CNN representations are quite good at clustering (blue curve in Fig.~\ref{fig:validation_losses}) when the number of characters is known.
However, using a threshold on validation to choose an operating point results in much lower performance (76 clusters instead of 13).
To further test this hypothesis, we train an MLP $\varphi_\theta$ to classify among our training set of actors, and use activations from the last layer as embeddings.
The orange curve in Fig.~\ref{fig:validation_losses} is lower than the base model (blue) indicating that training with more characters may have helped the base model.
Nevertheless, choosing an operating point is difficult.
Results in Table~\ref{table:test_episodes} rows 7-9 show that the CE over-clusters (create many more clusters than GT).
Directly using base CNN representations also results in many clusters (see supplementary).

\vspace{-5mm}
\paragraph{Verification losses.}
Next, we analyze LDML, contrastive, and triplet losses (Table~\ref{table:test_episodes} rows 10-18).
While these losses are often used to perform clustering, they are not designed for it \cite{wang2019centroidbased}.
We see two major features:
(i) unlike BCL, estimating the number of clusters is not a built-in feature and requires choosing a threshold on the validation set that may be unreliable; and
(ii) early errors in the iterative merging can really harm the overall composition.
We observe that triplet loss consistently achieves higher NMI and better estimates for number of clusters than contrastive and LDML.

\vspace{-5mm}
\paragraph{Prototypical loss (PRO) vs. BCL.}
Similar to verification losses (above), PRO works best when the number of clusters is known (\eg for few-shot learning). 
The loss has strong ties with $K$-means, and optimizes the space to create well separated $K$ clusters~\cite{law2018dimensionality}.
Interestingly, in our experiments, $\ell_2$ normalizing embeddings reduced the performance of PRO by over 15\% NMI.
We report PRO scores for non-normalized representations,
that are also more stable when transferring a threshold based on the validation set.
In fact, by comparing Table~\ref{table:test_episodes} row 19 with row 1, we see that PRO over-estimates the number of clusters when there are few background characters (BCL, row 22, works well here), but performs better in episodes with several background characters.

While PRO estimates more clusters, that does not translate to better assignments.
For example, on BUFFY-S5E4, PRO predicts 13 more clusters than BCL (35 vs. 22) and is closer to the ground-truth 32 clusters, but attains 7.7\% lower NMI and 3\% lower WCP.
A lower purity while having more clusters is a strong indicator of bad clustering.

We also compare performance between PRO and BCL when the number of clusters $K$ is known (see Table~\ref{table:test_known_K}).
BCL is able to consistently outperform both $K$-means or HAC clustering methods for the prototypical loss.
We also tried extensions of $K$-means that automatically determine the number of clusters in an unsupervised way~\cite{hamerly2004learning,pelleg2000x} when the representations are fixed.
Their performance was worse than our method of choosing a threshold on the validation set.
Additional comparisons are in the supplementary material.

\begin{figure}[t]
\centering
\vspace{-2mm}
\includegraphics[width=0.9\linewidth]{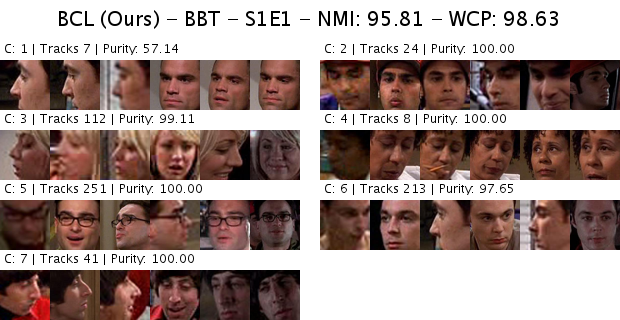} \vspace{2mm}
\includegraphics[width=0.9\linewidth]{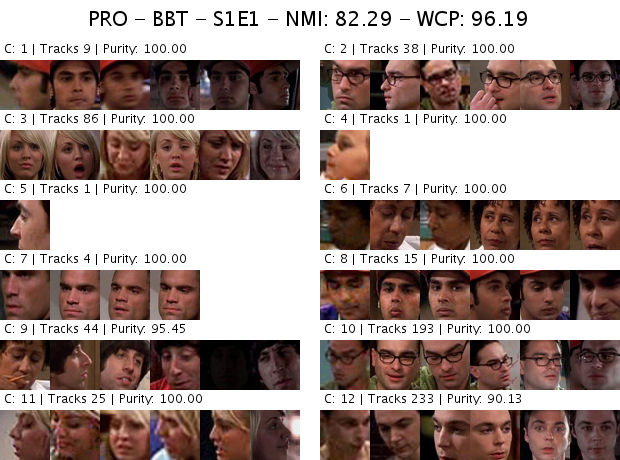}
\vspace{-3mm}
\caption{Visualizing clusters created by BCL (top) and PRO (bottom) for BBT-S1E1.
Refer to supp. material for other episodes.}
\vspace{-6mm}
\label{fig:qualitative}
\end{figure}

\vspace{-5mm}
\paragraph{Qualitative.}
Fig.~\ref{fig:qualitative} visualizes clusters created by BCL (top) vs. those with PRO (bottom) on BBT-S1E1.
BCL predicts 7 clusters in comparison to the ground-truth 8, and merges the singleton track of a background girl (C4 in PRO) with Penny (C3 in BCL).
Both methods find the other unnamed character - C4 in BCL, C6 in PRO.
While BCL merges few tracks of Sheldon and Kurt (C1), PRO is able to find Kurt (C7).
However, PRO splits clusters for Raj (C1, C8), Penny (C3, C11), and Leonard (C2, C10).

\vspace{-5mm}
\paragraph{Fine-tuning on each episode.}
Our model can be applied directly to several different datasets by using the learned threshold $4b$ without fine-tuning, this is a major advantage.
Following previous work that uses positive and negative pairs obtained automatically from each episode~\cite{cinbis2011unsupervised,tapaswi2014facecluster,zhang2016imptriplet}, BCL can be easily modified to fine-tune our model and make it cast-specific.
Shots with background characters are often crowded (multiple faces), and negative constraints among them can help resolve confusion.
Table~\ref{table:test_episodes} rows 25-27 show the overall performance improves after fine-tuning; importantly, the estimated number of characters (row 25) is much closer to the ground-truth (row 1).
Details of the fine-tuning procedure and comparison against fine-tuned baselines is in the supplementary material.





\vspace{-1mm}
\section{Conclusion}
\label{sec:conclusion}
\vspace{-1mm}
We presented \emph{Ball Cluster Learning} - a supervised approach to carve the representation space into balls of an equal radius.
We showed how the radius is related to the stopping criterion used in agglomerative clustering methods, and evaluated this approach for clustering face tracks in videos.
In particular, we considered a realistic setup where the number of clusters is not known, and tracks from all characters (main or otherwise) are included.
We reviewed several metric learning approaches and adapted them to this clustering setup.
BCL shows promising results, and to the best of our knowledge is the first approach that learns a threshold that can be used directly to estimate the number of clusters.

\vspace{-3mm}
\paragraph{Acknowledgments.}
This work was supported by the MSR-Inria joint lab, the Louis-Vuitton - ENS Chair on AI, DARPA Explainable AI (XAI), NSERC, and Samsung.


\balance
{\small
\bibliographystyle{ieee_fullname}
\bibliography{main}
}
\clearpage
\newpage

\appendix
\section*{Supplementary Material}

In this document, we discuss how we fine-tune our models on the test episodes using the BCL loss.
We also briefly discuss the challenges to fine-tune using triplet or prototypical losses, but show that they can use the constraints inspired by BCL in Sec.~\ref{sec:finetune}.
Additionally, we will show that variants of K-means that aim to predict the number of clusters such as X-means~\cite{pelleg2000x} and G-means~\cite{hamerly2004learning} perform worse than our proposed method (Sec.~\ref{sec:kmeans_predK}).
Finally, we present additional quantitative and qualitative results on the TV series episodes introduced in the main paper in Sec.~\ref{sec:additional_eval}.

\section{Fine-tuning on test episodes}
\label{sec:finetune}
As discussed in the related work section, many clustering approaches use unsupervised constraints that arise from the video to learn cast-specific metrics~\cite{cinbis2011unsupervised,tapaswi2014facecluster,wu2013simultaneous,wu2013constrained,xiao2014weighted,zhang2016imptriplet,zhang2016jfac}.
The positive constraints are obtained from face images within a track that are considered similar; and negative constraints from faces that appear at the same time in the video that can be assumed to be dissimilar.
Note that most previous works know the number of clusters, and use the constraints to improve the distance metric.

In the following, we show how our method can be modified to work with such positive and negative constraints.
We also discuss the limitations of the baselines for fine-tuning, but propose an alternative that uses BCL pair-wise constraints.

\paragraph{Ball Cluster Learning.}
Recall that our constraints are originally based on cluster samples and their centroids.
For all $x_i \in \Ccal_k$, BCL aims to satisfy $\dsf^2(\fvec_i, \mu_k) < b$ and $\dsf^2(\fvec_i, \mu_v) > \gamma$, where $\gamma = 9b + \epsilon$ and $x_i \not\in \Ccal_v$.

For a positive pair $x_i \in \Ccal_k, x_j \in \Ccal_k$, and a negative counterpart $x_u \in \Ccal_v$, the centroid constraints can be modified as:
\begin{equation} \label{eq:finetune_constraints}
\dsf^2(\fvec_i, \fvec_j) < 4b \,\, \mbox{ and } \,\, \dsf^2(\fvec_i, \fvec_u) > 4b + \epsilon .
\end{equation}

In practice, as the model is already trained, we wish to only fine-tune on the positive and negative pairs.
In most cases, the positive constraints are already satisfied, and using the relaxed constraint hurt performance.
Thus, we use: $\dsf^2(\fvec_i, \fvec_j) < \min \left( \dsf^2_{\mathrm{ori}}(\fvec_i, \fvec_j), 4b \right)$, where $\dsf^2_{\mathrm{ori}}(\fvec_i, \fvec_j)$ is the distance between the pair prior to fine-tuning.
The constraints are formulated as loss functions to train the model by using the $[\cdot]_+ = \max(0, \cdot)$ operator as before. 

We tried an analogous strategy for dissimilar pairs (using $\max \left( \dsf^2_{\mathrm{ori}}(\fvec_i, \fvec_u), 4b + \epsilon \right)$) but it did not provide significant performance improvement.
Thus, we ignored it.

During training, we freeze the ball squared-radius $b$, use a learning rate of 0.0003 (0.1 times the original), and select a random face image from each track in the pairs (about 1,000 pairs for each episode).
The model parameters are updated for a fixed number of iterations (2,000 for all single episodes).
As before, we perform clustering by using Hierarchical Agglomerative Clustering (HAC) with complete linkage and distance threshold $\tau = 4b$.

\paragraph{Triplet loss~\cite{schroff2015facenet}}
is well suited to train a model with above-mentioned automatically obtained positive and negative pairs.
However, it does not involve learning a threshold that can be used directly with HAC (or any clustering).
As we are fine-tuning on test episodes, we do not have access to a validation set that would allow us to obtain such a threshold.
Furthermore, optimizing performance on each test episode by choosing a new \emph{best} threshold is inappropriate.

To circumvent this, we use the original threshold learned on the validation set $\tau$ and formulate pair-wise constraints in a similar manner to BCL.
In particular, the positive pairs follow $\dsf^2(\fvec_i, \fvec_j) < \tau$ and negative pairs $\dsf^2(\fvec_i, \fvec_u) > \tau$.
Thus, we fine-tune the model checkpoint trained using the triplet loss with the BCL loss.
All other implementation details (learning rate, number of iterations, \etc) are same as those used for BCL.

\paragraph{Prototypical loss~\cite{snell2017prototypical}.}
Unlike the triplet loss that is designed to work with samples (triplets), the prototypical loss needs class/cluster centroids.
It also faces the same challenge of not knowing the number of clusters, or not having a clear stopping criterion for HAC.
Nevertheless, as discussed for triplet loss above, we adopt BCL pair-wise constraints for the prototypical loss with the threshold $\tau$ chosen on validation.
All other details are kept same.

\begin{table*}[h]
\centering
\footnotesize
\tabcolsep=0.17cm
\begin{tabular}{rl|cccccc|cccccc|ccc}
\toprule
    & & \multicolumn{6}{c|}{BBT}  & \multicolumn{6}{c|}{BUFFY} & BBT & BUFFY & BOTH \\
    & &  S1E1  &  S1E2  &  S1E3  &  S1E4  &  S1E5  &  S1E6  &  S5E1  &  S5E2  &  S5E3  &  S5E4  &  S5E5  &  S5E6  & 6 ep. & 6 ep. & 12 ep.  \\
\midrule
 1 & $\#$Ch      &   8 &   6 &  26 &  28 &  25 &  37 &  13 &  22 &   15 &  32 &  38 &   45 &  103  &   109  &   212  \\
\hline
\multicolumn{17}{>{\columncolor{mylightorange}}c}{Pre-trained Triplet Loss~\cite{schroff2015facenet} + BCL Fine-tune} \\[0.1mm]
 2 & $\#$Cl      &  6   &   7   &   13  &   12  &   12  &   23  &   16  &   21  &   16  &   18  &   22  &   23  &   33  &   54  &   73  \\
 3 & NMI         & \textbf{97.98} & 97.13 & 91.22 & 86.37 & 95.25 & 83.38 & \textbf{85.91} & 82.62 & 78.98 & 75.34 & 76.04 & 78.66 & \textbf{89.09} & 76.28 & 78.95 \\
 4 & WCP         & 99.09 & \textbf{99.84} & 92.73 & 89.23 & 94.08 & 86.31 & 91.95 & 87.71 & 85.76 & 78.95 & 80.83 & 79.59 & 92.40 & 81.45 & 82.26 \\
\hline
\multicolumn{17}{>{\columncolor{mygreen}}c}{Pre-trained Prototypical Loss~\cite{snell2017prototypical} + BCL Fine-tune} \\[0.1mm]
 5 & $\#$Cl      &  6    &  \textbf{6}    &  19   &  16   &  15   &  41   &  17   &  22   &  21   &  27   &  21   &  34   &   61  &   72  &  113   \\
 6 & NMI         & 96.76 & 97.09 & 91.43 & \textbf{90.83} & \textbf{95.84} & \textbf{85.38} & 77.98 & \textbf{83.01} & \textbf{79.29} & 77.24 & \textbf{81.74} & \textbf{82.31} & 87.74 & \textbf{79.90} & \textbf{82.81} \\
 7 & WCP         & 98.17 & \textbf{99.67} & 95.00 & 93.31 & 94.85 & \textbf{93.69} & 90.82 & 87.11 & 88.86 & 82.52 & 84.76 & 84.80 & 93.96 & 85.17 & 85.26 \\
\hline
\multicolumn{17}{>{\columncolor{mycyan}}c}{Pre-trained Ball Cluster Learning (Ours) + BCL Fine-tune} \\[0.1mm]
 8 & $\#$Cl      &   \textbf{9}   &   8   &   \textbf{24}  &   \textbf{24}  &   \textbf{21}  &   \textbf{36}  &   23  &   27  &   25  &   \textbf{36}  &   \textbf{38}  &   \textbf{40}  &  \textbf{69}   &  \textbf{78}   &  \textbf{126}  \\
 9 & NMI         & 97.34 & \textbf{97.80} & \textbf{94.00} & 90.42 & \textbf{95.83} & 83.32 & 84.59 & 82.59 & 78.76 & \textbf{77.58} & \textbf{81.71} & 79.51 & 88.26 & 77.05 & 80.42 \\
10 & WCP         & \textbf{99.24} & \textbf{99.67} & \textbf{96.06} & \textbf{96.08} & \textbf{97.71} & 90.36 & \textbf{94.97} & \textbf{88.12} & \textbf{90.28} & \textbf{86.19} & \textbf{90.24} & \textbf{88.13} & \textbf{94.11} & \textbf{86.64} & \textbf{85.84} \\
\bottomrule
\end{tabular}
\vspace{1mm}
\caption{Clustering performance on episodes of the test set, with fine-tuning on the test set.
The last three columns show results on datasets created by combining tracks from several episodes.
$\#$Ch is the ground-truth number of characters (row 1); and
$\#$Cl is number of predicted clusters and should be close to the number of characters.
Read this table by looking at each column, and seeing which method is able to predict the number of clusters and has high NMI and WCP scores.
}
\label{table:finetuned_episodes}
\end{table*}

\paragraph{Evaluation.}
Table~\ref{table:finetuned_episodes} presents results of fine-tuning on each episode.
For each episode, we see that BCL fine-tuned model (BCL-FT) is able to predict the number of clusters quite accurately.
We believe this can be attributed to background characters often appearing simultaneously, providing sufficient negative constraints.
On the combined episodes datasets (last three columns of Table~\ref{table:finetuned_episodes}), the negative constraints are within each episode, and it is not possible to distinguish between background characters across episodes.
This explains why BCL predicts fewer clusters than ground-truth.

Both baselines (TRI-FT and PRO-FT) also show good performance improvements after fine-tuning with BCL.
With respect to predicting the number of clusters, PRO-FT and BCL-FT seem to have flipped roles, with PRO-FT now predicting fewer clusters after fine-tuning, but over-clustering before fine-tuning (see Table \ref{table:test_episodes} of the main paper).
However, note that BCL-FT with more clusters has higher purity, while PRO (no fine-tune) had lower purity even with more clusters (in Table \ref{table:test_episodes}).

\paragraph{Conclusion.}
This experiment emphasizes that BCL combines the best of all worlds.
Models can be trained with both samples and centroids, or pairs and triplets.
Most importantly, BCL learns a threshold to predict the number of clusters automatically.
In addition, BCL pair-wise loss can be used to fine-tune models trained with other losses to achieve performance gains.

\section{K-means variants}
\label{sec:kmeans_predK}

\begin{table*}[h]
\centering
\footnotesize
\tabcolsep=0.17cm
\begin{tabular}{rl|cccccc|cccccc|ccc}
\toprule
    & & \multicolumn{6}{c|}{BBT}  & \multicolumn{6}{c|}{BUFFY} & BBT & BUFFY & BOTH \\
    & &  S1E1  &  S1E2  &  S1E3  &  S1E4  &  S1E5  &  S1E6  &  S5E1  &  S5E2  &  S5E3  &  S5E4  &  S5E5  &  S5E6  & 6 ep. & 6 ep. & 12 ep.  \\
\midrule
 1 & $\#$Ch      &   8 &   6 &  26 &  28 &  25 &  37 &  13 &  22 &   15 &  32 &  38 &   45 &  103  &   109  &   212  \\
\midrule
\multicolumn{17}{c}{X-Means on Base CNN representation} \\
 2 & $\#$Cl      &   41  &   42  &   45  &   41  &   51  &   48  &   93  &   89  &   90  &   82  &  101  &   91  &  175  &  180  &  327  \\
 3 & NMI         & 55.74 & 56.83 & 65.75 & 64.89 & 66.47 & 66.51 & 59.61 & 66.86 & 57.56 & 62.99 & 61.03 & 66.23 & 56.26 & 62.67 & 65.85 \\
 4 & WCP         & 98.63 & 97.89 & 92.73 & 90.70 & 94.27 & 84.64 & 94.59 & 90.74 & 88.86 & 87.75 & 88.21 & 86.15 & 92.71 & 87.21 & 88.68 \\
\multicolumn{17}{c}{X-Means on features learned with BCL} \\
 5 & $\#$Cl      &   42  &   25  &   45  &   34  &   43  &   44  &   97  &   41  &   81  &   84  &   80  &   91  &  162  &  163  &  313  \\
 6 & NMI         & 56.90 & 62.46 & 69.82 & 68.35 & 71.46 & 70.08 & 60.12 & 72.16 & 59.99 & 65.90 & 66.59 & 69.72 & 58.62 & 63.91 & 66.63 \\
 7 & WCP         & 99.24 & 97.89 & 95.45 & 91.35 & 96.18 & 87.38 & 95.85 & 88.62 & 91.04 & 91.31 & 92.02 & 89.48 & 93.22 & 88.73 & 88.87 \\
\midrule
\midrule
\multicolumn{17}{c}{G-Means on Base CNN representation} \\
 8 & $\#$Cl      &   57  &   31  &   56  &   33  &   46  &   67  &   87  &  101  &  117  &   74  &   73  &  101  &  243  &  404  &  567  \\
 9 & NMI         & 55.29 & 62.61 & 70.41 & 71.48 & 71.74 & 67.00 & 58.07 & 66.63 & 58.34 & 67.06 & 63.50 & 69.08 & 57.64 & 61.57 & 66.07 \\
10 & WCP         & 98.32 & 96.42 & 94.85 & 91.35 & 94.85 & 86.55 & 89.18 & 89.93 & 88.94 & 89.31 & 87.14 & 87.68 & 92.14 & 87.21 & 87.95 \\
\multicolumn{17}{c}{G-Means on features learned with BCL} \\
11 & $\#$Cl      &   23  &   39  &   41  &   35  &   45  &   79  &   42  &   54  &   75  &   63  &   72  &  106  &  137  &  337  &  481  \\
12 & NMI         & 68.51 & 64.54 & 72.74 & 70.38 & 73.22 & 66.18 & 67.87 & 61.43 & 62.88 & 65.70 & 66.91 & 69.09 & 60.86 & 61.63 & 64.42 \\
13 & WCP         & 98.93 & 98.05 & 92.42 & 91.35 & 94.85 & 87.02 & 91.07 & 71.30 & 84.00 & 86.41 & 88.45 & 87.86 & 89.94 & 85.51 & 84.06 \\
\bottomrule
\end{tabular}
\vspace{1mm}
\caption{Clustering performance on episodes of the test set using two variants of K-means that predict the number of clusters.}
\label{table:kmeans_variants}
\end{table*}

Owing to the popularity of the $K$-means approach for clustering, there is some work on automatically estimating the number of clusters while performing clustering based on some criterion.
In this section, we will look at two such methods and analyze how they work when applied to our challenging datasets.
Note that both these methods do not further learn an embedding, but rely on existing features.
Thus, we evaluate performance on the Base CNN representations -- that are actually quite good (see Table~\ref{table:base_representation}), as well as the features learned using our BCL loss function.

\paragraph{X-means~\cite{pelleg2000x}}
In this variant, clustering starts with all samples in the same cluster and splits until some stopping criterion.
In particular, at each iteration, a cluster is split into two components.
The Bayesian Information Criterion (BIC) is used to decide whether the newly created two clusters are preferred over the original single cluster.
Clustering stops when a maximum number of clusters $K_{\max}$ has been crossed, or when further splitting any cluster would result in lowered BIC scores.

Table~\ref{table:kmeans_variants} rows 2-4 show the performance of X-means when using base CNN features, and rows 5-7 when using features trained with BCL loss.
We choose $K_{\max}$ to be 40 for the BBT episodes, 80 for BUFFY, 150 for BBT (6 episodes combined) and BUFFY (6 episodes combined), and 300 for BOTH (all 12 episodes).
These are strong upper bounds for all datasets.
However, as seen from the results, the method stops the iterations only after crossing the maximum number of clusters in all datasets (\ie~all predicted number of clusters are higher than $K_{\max}$).
This, together with the poor NMI scores, suggests that the using BIC may not be a sufficiently strong criterion for a complex dataset.

\paragraph{G-means~\cite{hamerly2004learning}}
Similar to X-means, this approach also starts with all samples in one cluster, and iteratively splits them based on some criterion.
Different to X-means, the stopping criterion used here determines the ``Gaussian-ness'' of samples around the cluster centroid.
In particular, clusters that have a strong Gaussian shape are not split further, while others (\eg~those that may be bimodal) are split into two.
This process repeats until no more clusters can be split.
The Anderson-Darling test is used to determine whether a distribution is Gaussian.

We present the results of G-means in the second half of Table~\ref{table:kmeans_variants}.
We see that G-means also fails at reliably estimating the number of clusters, and prefers to overcluster all datasets.

\section{Additional evaluation}
\label{sec:additional_eval}
\paragraph{Base CNN representations.}

\begin{table*}[h]
\centering
\footnotesize
\tabcolsep=0.17cm
\begin{tabular}{rl|cccccc|cccccc|ccc}
\toprule
    & & \multicolumn{6}{c|}{BBT}  & \multicolumn{6}{c|}{BUFFY} & BBT & BUFFY & BOTH \\
    & &  S1E1  &  S1E2  &  S1E3  &  S1E4  &  S1E5  &  S1E6  &  S5E1  &  S5E2  &  S5E3  &  S5E4  &  S5E5  &  S5E6  & 6 ep. & 6 ep. & 12 ep.  \\
\midrule
 1 & $\#$Ch      &   8 &   6 &  26 &  28 &  25 &  37 &  13 &  22 &   15 &  32 &  38 &   45 &  103  &   109  &   212  \\
\midrule
\multicolumn{17}{c}{Base CNN representation} \\
 2 & $\#$Cl      &  36   &  38   &  49   &  51   &  36   &  59   &  76   &  75   &  94   &  93   &  95   &  92   &  200  &  407  &  609  \\
 3 & NMI         & 59.84 & 60.60 & 68.18 & 69.18 & 74.85 & 68.42 & 61.66 & 67.94 & 57.88 & 65.01 & 65.59 & 66.90 & 59.33 & 60.00 & 64.80 \\
 4 & WCP         & 97.41 & 97.72 & 92.88 & 92.66 & 95.42 & 86.31 & 91.95 & 88.02 & 88.27 & 89.64 & 91.31 & 85.52 & 93.19 & 88.08 & 89.14 \\
\bottomrule
\end{tabular}
\vspace{1mm}
\caption{Clustering performance on episodes of the test set when using base CNN representation.}
\label{table:base_representation}
\end{table*}

In Table~\ref{table:base_representation}, we present the performance of base CNN representations with standard HAC clustering and using a threshold learned on the validation set.
These results should be analyzed together with Table \ref{table:test_episodes} of the main paper, but were omitted due to space constraints.
Note that the threshold is chosen such that correct number of clusters are created on the validation set.
While the base representation is quite good (as seen in NMI and WCP curves), choosing a threshold is an unreliable method and results in over-clustering on the test episodes.

\paragraph{When $K$ is known.}

\begin{table*}
\centering
\footnotesize
\tabcolsep=0.15cm
\begin{tabular}{rl|cccccc|cccccc|ccc}
\toprule
\multirow{2}{*}{Method} & \multirow{2}{*}{Metric} & \multicolumn{6}{c|}{BBT}  & \multicolumn{6}{c|}{BUFFY} & BBT & BUFFY & ALL \\
& &  S1E1  &  S1E2  &  S1E3  &  S1E4  &  S1E5  &  S1E6  &  S5E1  &  S5E2  &  S5E3  &  S5E4  &  S5E5  &  S5E6  & 6 ep. & 6 ep. & 12 ep.  \\
\hline
\multicolumn{17}{>{\columncolor{mylightorange}}c}{Triplet Loss~\cite{schroff2015facenet}} \\[0.1mm]
KM  & NMI & 73.2 & 83.4 & 74.7 & 70.5 & 75.5 & 67.8 & 74.3 & 64.7 & 68.7 & 64.6 & 68.8 & 66.0 & 57.8 & 59.2 & 62.6 \\
KM  & WCP & 93.6 & 96.7 & 93.2 & 91.4 & 93.7 & 83.7 & 88.1 & 72.3 & 80.8 & 80.1 & 86.1 & 78.0 & 89.7 & 77.7 & 80.5 \\
HAC & NMI & 88.3 & 69.0 & 79.1 & 76.6 & 81.9 & 70.8 & 76.2 & 64.1 & 66.9 & 64.9 & 70.7 & 67.6 & 64.0 & 60.1 & 65.3 \\
HAC & WCP & \textbf{98.5} & 79.2 & 93.0 & 91.5 & 94.7 & 80.4 & 86.4 & 67.4 & 77.3 & 73.3 & 84.0 & 76.5 & 90.2 & 72.7 & 76.9 \\
\hline
\multicolumn{17}{>{\columncolor{mygreen}}c}{Prototypical Loss~\cite{snell2017prototypical}} \\[0.1mm]
KM  & NMI & 80.8 & 82.5 & 76.0 & 72.3 & 76.0 & 69.4 & 79.5 & 74.6 & 74.3 & 71.9 & 71.4 & 72.5 & 60.7 & 64.5 & 66.8 \\
KM  & WCP & 95.0 & 95.6 & 93.6 & \textbf{91.7} & 94.3 & 83.7 & 89.4 & 84.2 & \textbf{83.7} & 86.0 & 88.9 & 84.7 & 91.3 & 85.4 & 85.6 \\
HAC & NMI & 87.6 & 86.6 & 83.1 & \textbf{80.3} & \textbf{89.2} & 74.0 & 67.9 & 68.4 & 77.4 & 70.6 & 76.5 & 73.8 & 68.3 & 65.8 & 70.3 \\
HAC & WCP & 94.7 & 94.8 & \textbf{94.2} & 91.0 & \textbf{96.2} & 85.2 & 74.8 & 73.0 & 79.9 & 81.2 & 85.1 & 82.8 & 91.1 & 80.0 & 83.3 \\
\hline
\multicolumn{17}{>{\columncolor{mycyan}}c}{Ball Cluster Learning (Ours)} \\[0.1mm]
KM  & NMI & 83.9 & 90.1 & 73.2 & 70.2 & 76.9 & 71.6 & 78.9 & \textbf{79.1} & 72.0 & 72.9 & 71.6 & 74.8 & 60.5 & 66.7 & 68.7 \\
KM  & WCP & \textbf{98.5} & 99.2 & 92.9 & 91.2 & 93.5 & 86.1 & \textbf{91.6} & \textbf{87.4} & 81.6 & \textbf{87.6} & 88.8 & 87.7 & 92.0 & \textbf{87.3} & \textbf{88.0} \\
HAC & NMI & \textbf{92.8} & \textbf{91.9} & \textbf{84.3} & 78.5 & 86.1 & \textbf{76.1} & \textbf{81.3} & 75.3 & \textbf{77.9} & \textbf{75.9} & \textbf{76.9} & \textbf{78.6} & \textbf{70.6} & \textbf{69.1} & \textbf{72.5} \\
HAC & WCP & \textbf{98.6} & \textbf{98.2} & 92.6 & \textbf{91.7} & 96.0 & \textbf{86.7} & 89.6 & 81.4 & 81.4 & 87.3 & \textbf{91.2} & \textbf{88.5} & \textbf{93.0} & 85.3 & 86.2 \\
\bottomrule
\end{tabular}
\vspace{0.1cm}
\caption{Comparison between models trained with triplet, prototypical, and our approach when the number of clusters is known.
We evaluate both $K$-means (KM) as well as Hierarchical Agglomerative Clustering (HAC) to obtain the appropriate number of clusters.
A short version of this appeared as Table \ref{table:test_known_K} in the main paper.}
\label{table:knownK_full_comparison}
\end{table*}

We also present results when (for some reason) the number of clusters $K$ is known.
We compare against best performing baselines: triplet and prototypical loss, on all episodes of the test set.
Table~\ref{table:knownK_full_comparison} shows that our method is able to achieve higher NMI and WCP in most cases (12 out of 15).
Note that this experiment is presented for completeness, as the main point of BCL is to automatically predict the number of clusters, when $K$ is unknown.
All results are without fine-tuning.

\paragraph{Choosing a threshold on train set.}

\begin{table*}
\centering
\footnotesize
\tabcolsep=0.09cm
\begin{tabular}{l|cccccc|cccccc}
\toprule
Thresh & \multicolumn{6}{c|}{BBT}  & \multicolumn{6}{c}{BUFFY} \\
Set    &  S1E1  &  S1E2  &  S1E3  &  S1E4  &  S1E5  &  S1E6  &  S5E1  &  S5E2  &  S5E3  &  S5E4  &  S5E5  &  S5E6  \\
\midrule
$\#$Ch   &   8 &   6 &  26 &  28 &  25 &  37 &  13 &  22 &   15 &  32 &  38 &   45 \\
\hline
\multicolumn{13}{>{\columncolor{mylightorange}}c}{Contrastive Loss~\cite{chopra2005learning}} \\[0.1mm]
train & 374 (40.6) & 382 (41.0) & 411 (52.5) & 443 (51.7) & 341 (54.3) & 629 (56.7) & 605 (48.1) & 721 (55.0) & 862 (46.8) & 694 (53.5) & 608 (53.7) & 725 (57.9) \\
val   &  14 (62.5) &  13 (63.7) &  17 (61.8) &  22 (65.6) &  19 (71.4) &  32 (55.7) &  22 (61.0) &  30 (53.9) &  26 (58.2) &  29 (53.4) &  29 (53.6) &  27 (52.0) \\
\hline
\multicolumn{13}{>{\columncolor{mylightorange}}c}{Triplet Loss~\cite{schroff2015facenet}} \\[0.1mm]
train & 28 (65.5) & 31 (63.4) & 38 (74.6) & 44 (74.1) & 39 (78.2) & 72 (68.3) & 64 (65.0) & 73 (64.9) & 77 (60.1) & 67 (63.6) & 66 (69.4) & 79 (68.4) \\
val   &  9 (88.1) & 12 (71.2) & 15 (79.8) & 16 (76.7) & 13 (85.8) & 23 (69.3) & 23 (73.6) & 24 (64.2) & 25 (66.2) & 22 (63.6) & 23 (67.9) & 26 (65.5) \\
\hline
\multicolumn{13}{>{\columncolor{mygreen}}c}{Prototypical Loss~\cite{snell2017prototypical}} \\[0.1mm]
train & 19 (74.6) & 25 (69.3) & 35 (77.4) & 39 (76.6) & 27 (86.6) & 63 (73.9) & 50 (70.7) & 55 (69.7) & 43 (67.8) & 60 (69.7) & 63 (72.7) & 65 (74.3) \\
val   & 12 (82.3) & 15 (75.1) & 22 (83.7) & 28 (80.3) & 18 (91.4) & 41 (74.3) & 32 (74.2) & 32 (71.0) & 20 (76.2) & 35 (70.5) & 40 (76.6) & 36 (73.5) \\
\bottomrule
\end{tabular}
\vspace{1mm}
\caption{Choosing the HAC threshold on train vs. validation set.
Showing the number of predicted clusters and NMI.
Ideal number of clusters is presented in the first row.
Note how it is beneficial to have a separate validation set, as overfitting on training can lead to selection of smaller thresholds.}
\label{table:threshold_on_train}
\end{table*}

As the training set is much larger than validation, it might seem that the baselines may perform better when choosing a threshold on the validation set.
However, this is not the case as observed from Table~\ref{table:threshold_on_train}.
As the MLP $\varphi_\theta$ fits well to the training set a smaller cutoff threshold (distance) is selected resulting in more clusters on unseen data.
Thus, it is important to have a separate validation set.

\begin{figure*}[t]
\begin{center}
\includegraphics[width=0.23\linewidth]{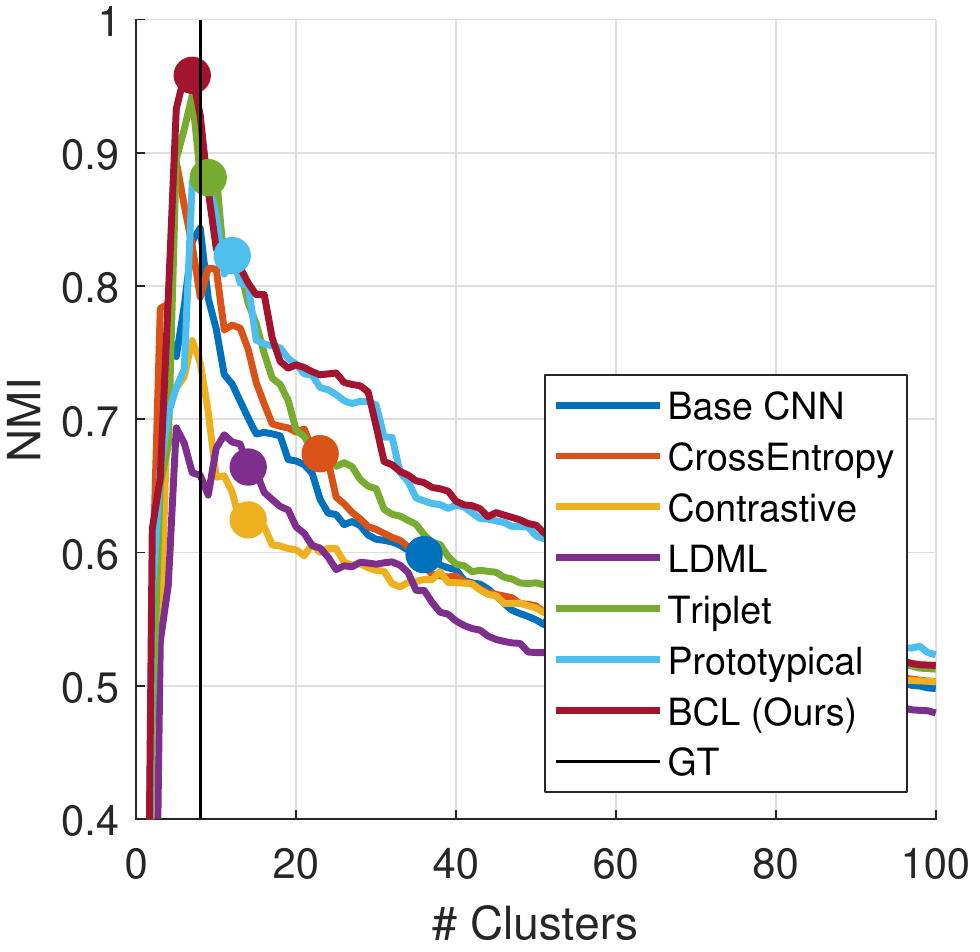}
\includegraphics[width=0.23\linewidth]{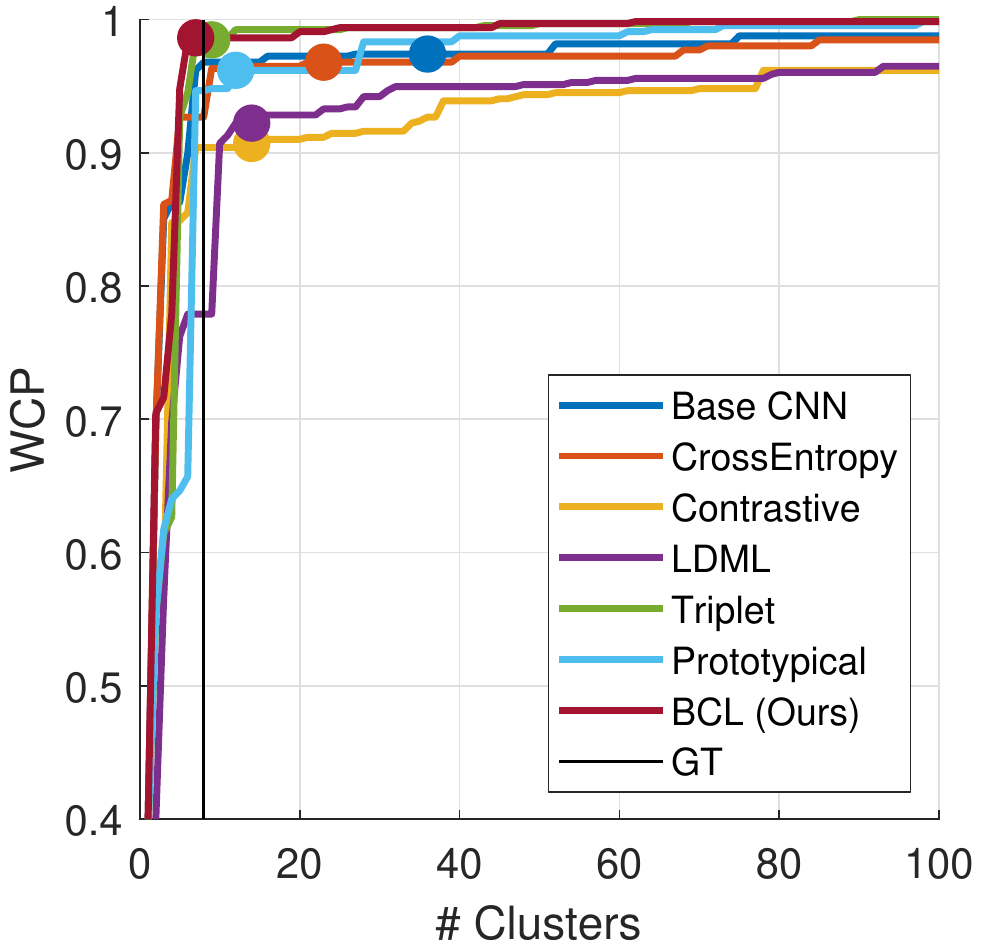} \quad\quad
\includegraphics[width=0.23\linewidth]{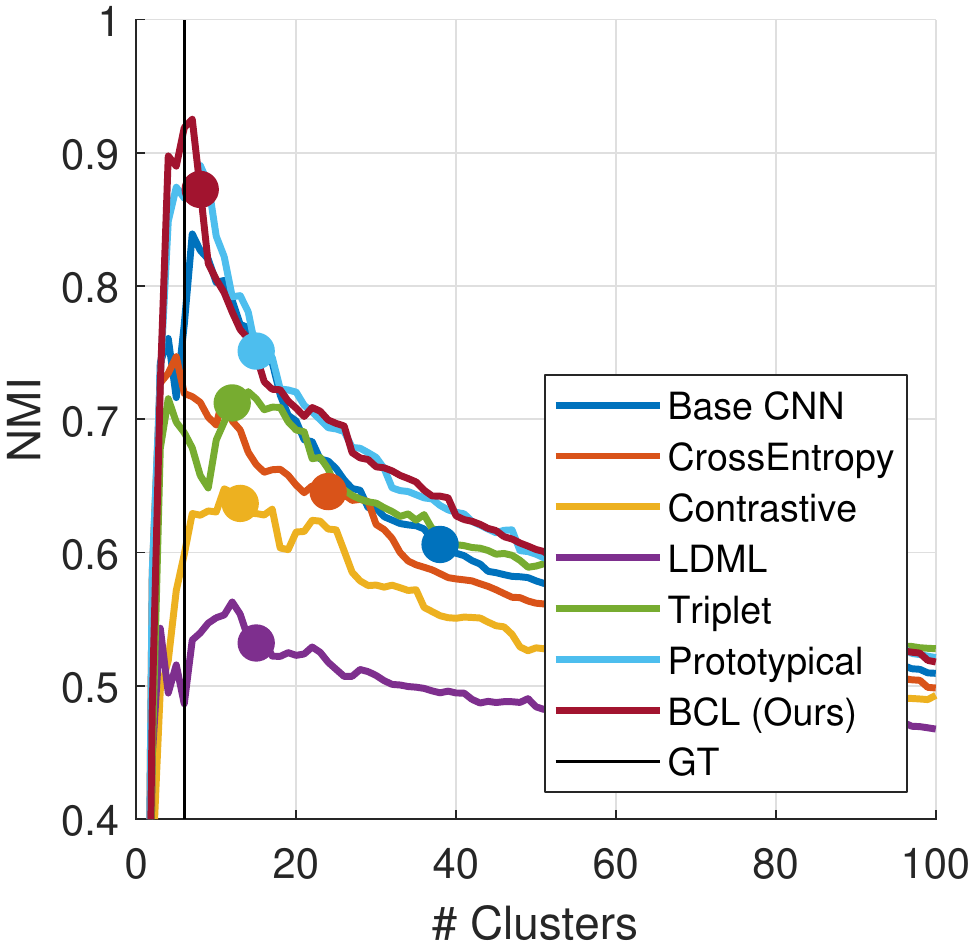}
\includegraphics[width=0.23\linewidth]{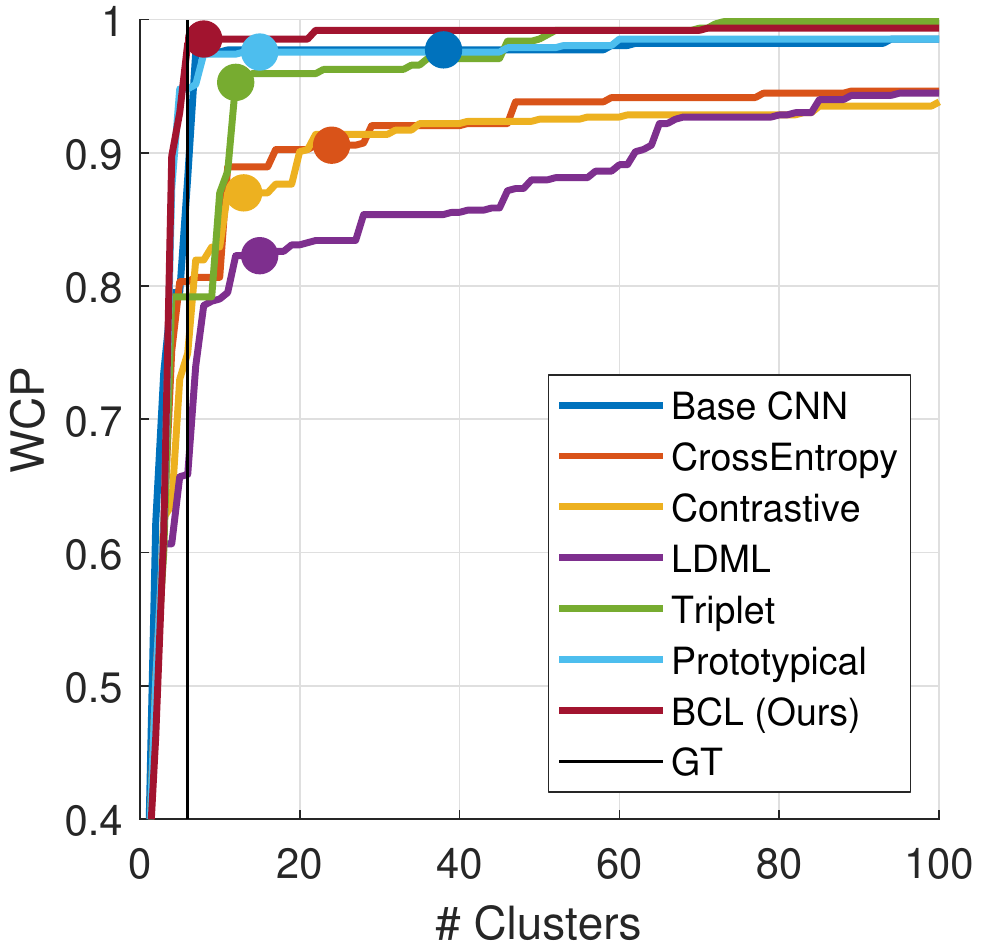} \\ \vspace{6mm}
\includegraphics[width=0.23\linewidth]{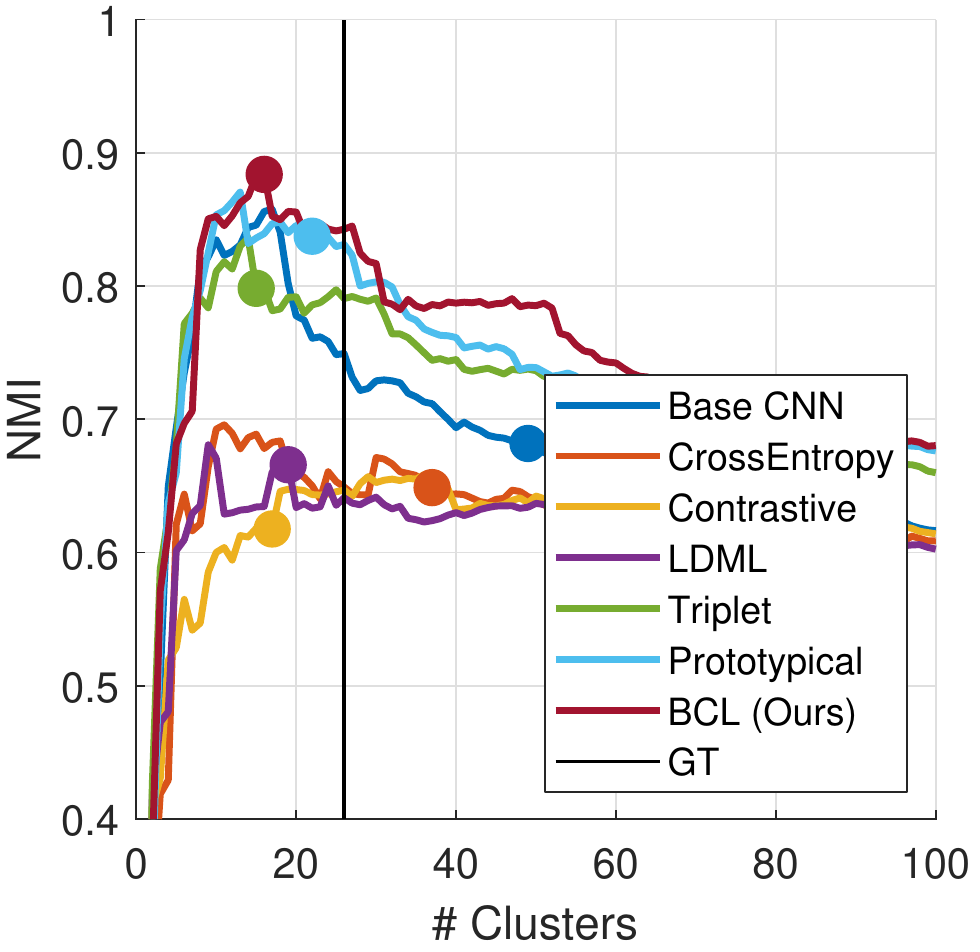}
\includegraphics[width=0.23\linewidth]{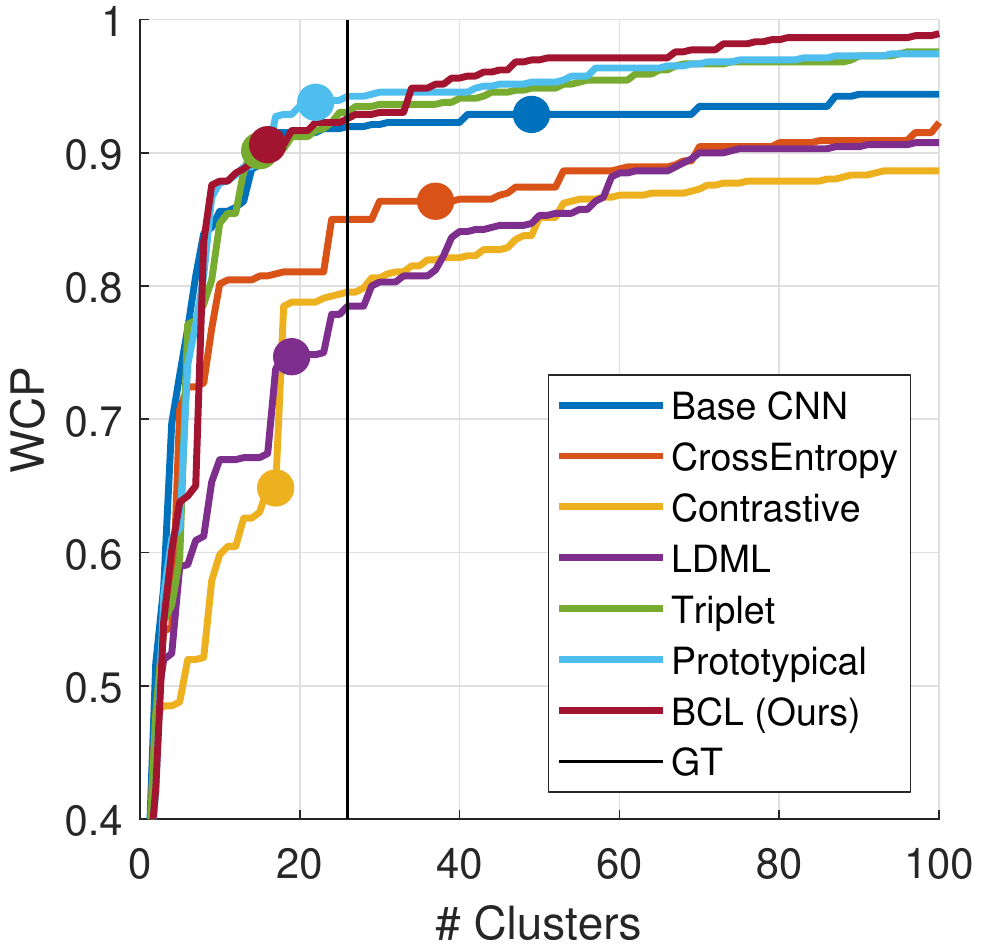} \quad\quad
\includegraphics[width=0.23\linewidth]{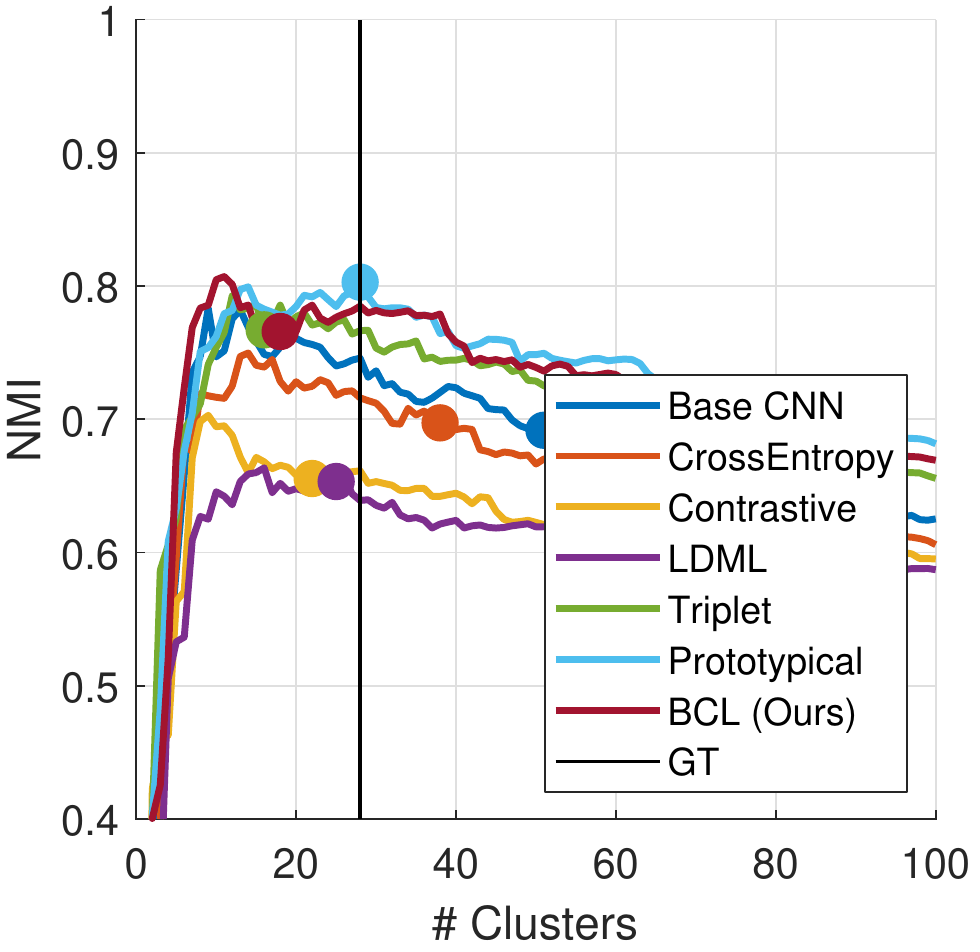}
\includegraphics[width=0.23\linewidth]{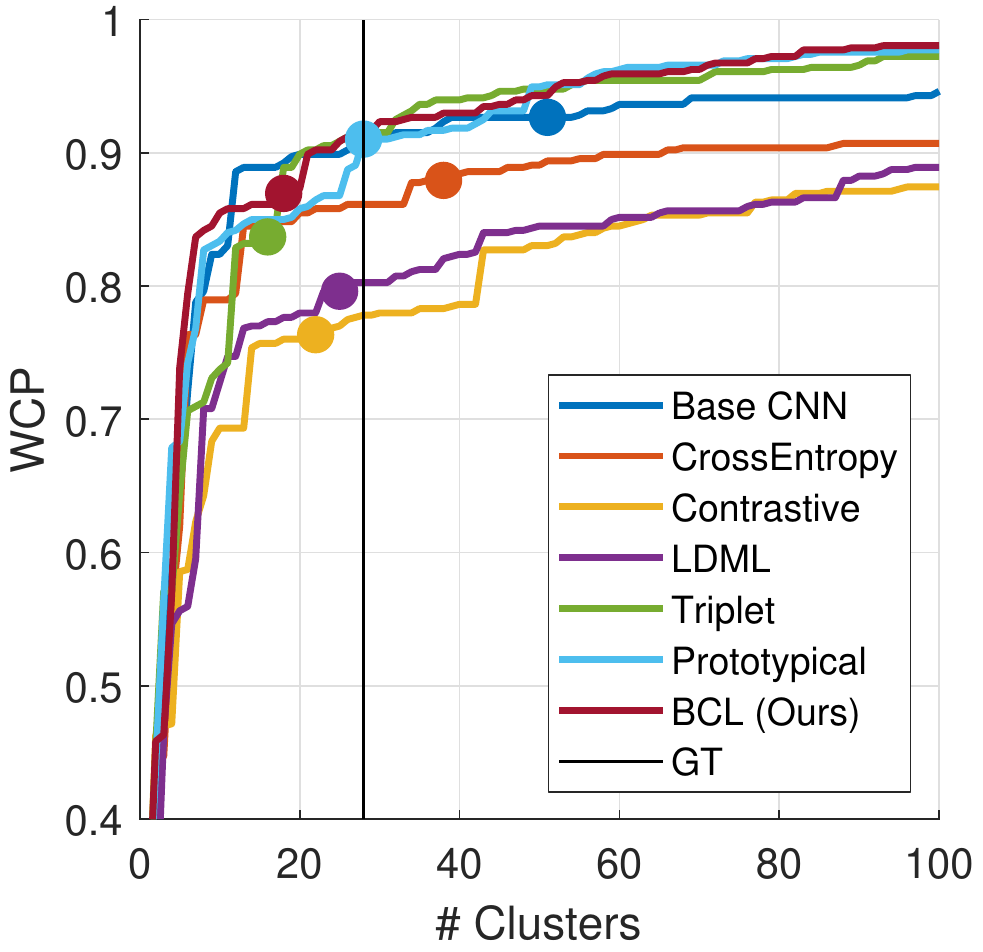} \\ \vspace{6mm}
\includegraphics[width=0.23\linewidth]{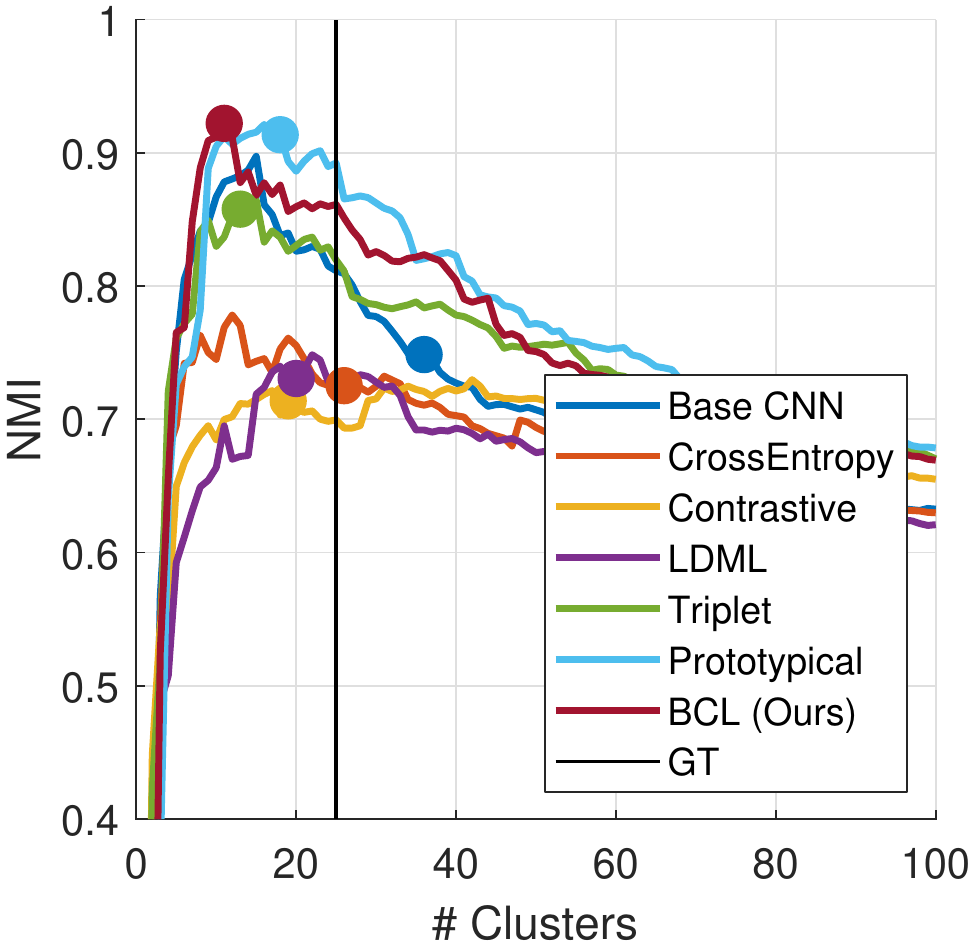}
\includegraphics[width=0.23\linewidth]{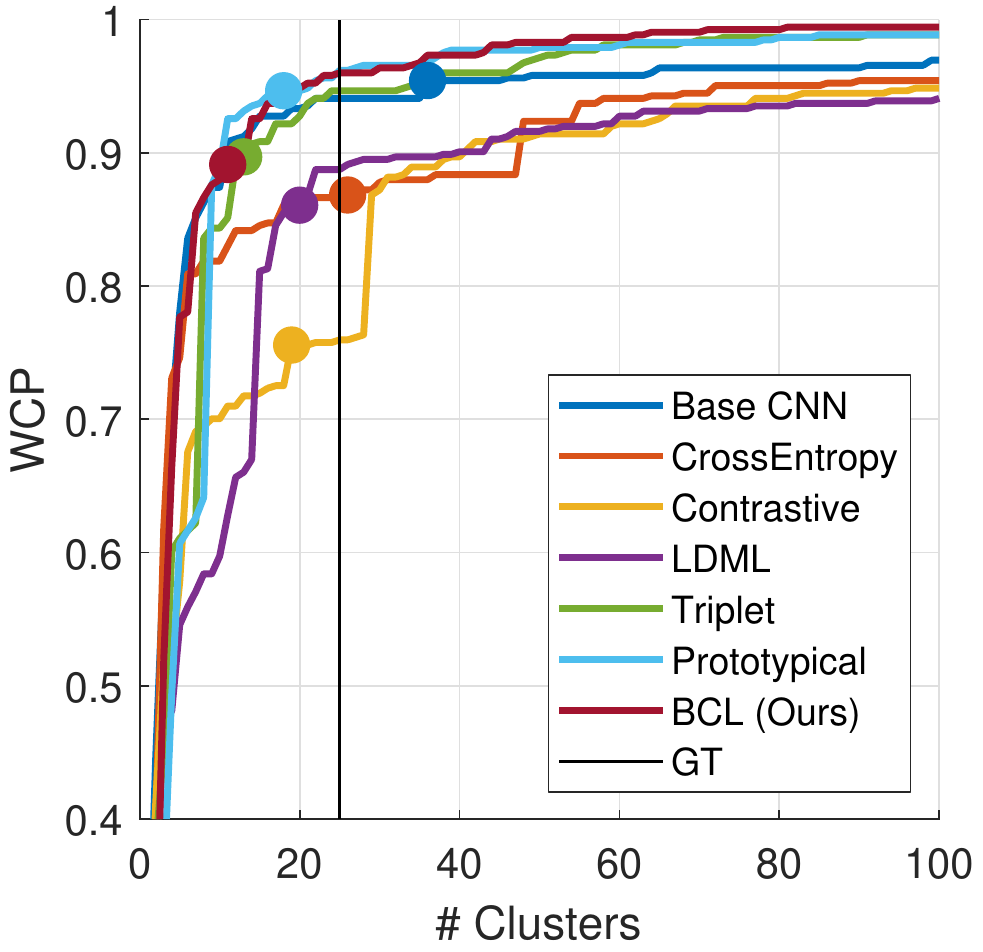} \quad\quad
\includegraphics[width=0.23\linewidth]{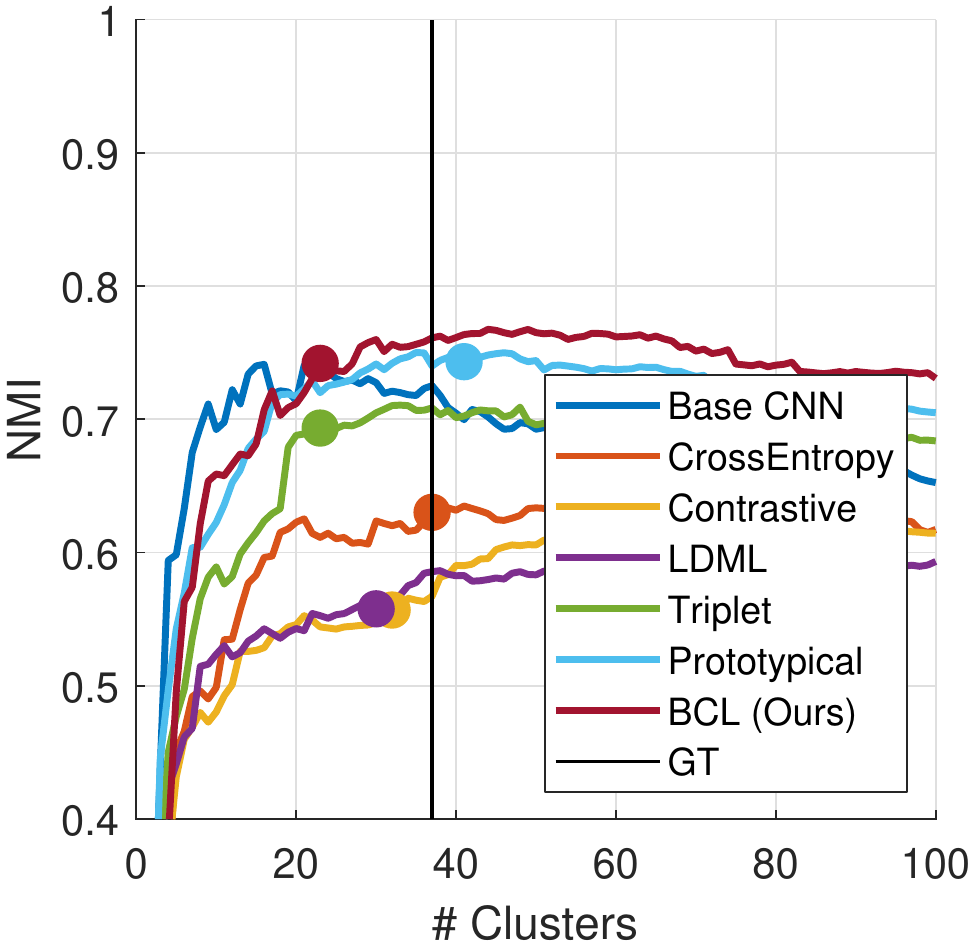}
\includegraphics[width=0.23\linewidth]{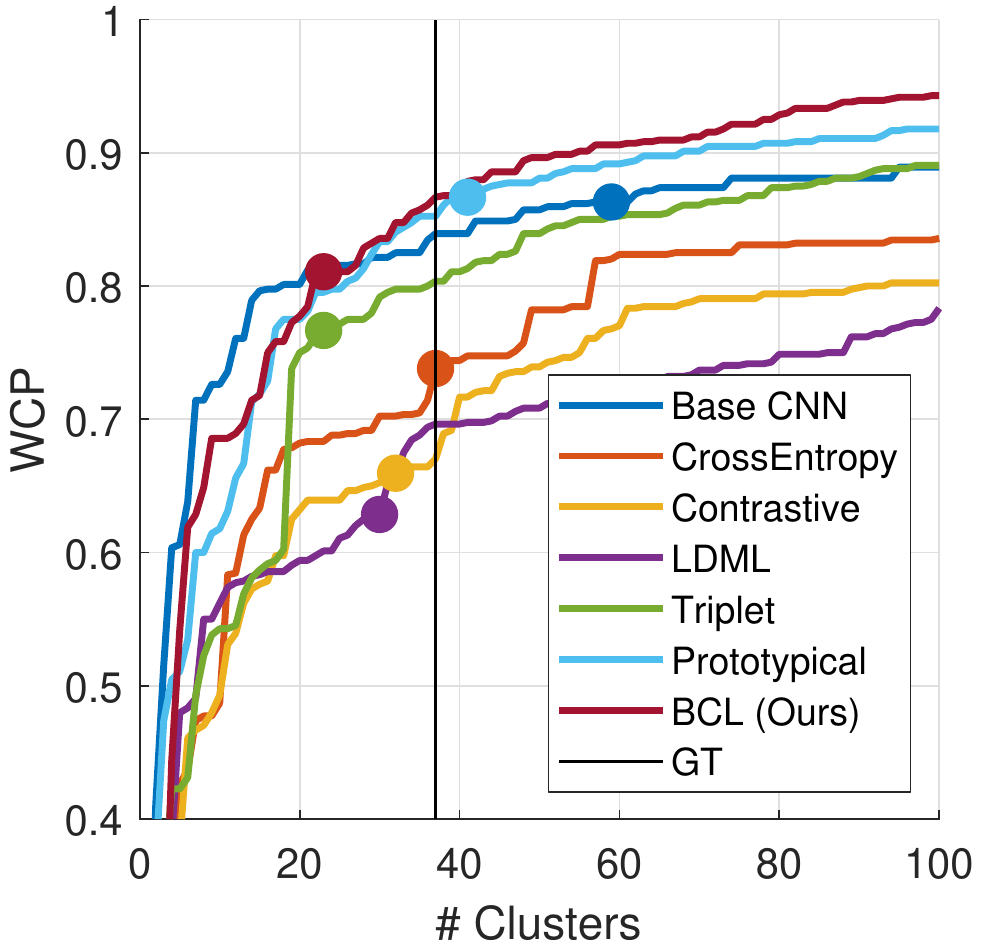}
\end{center}
\vspace{-2mm}
\caption{NMI and WCP vs. number of clusters for BBT-S1E1 to S1E6 (left to right, top to bottom).
Circles indicate operating points (\ie number of predicted clusters for the methods), our method uses the HAC threshold 4$b$, while all others are using the threshold tuned to give 66 clusters on the validation set.
Best seen in color.}
\vspace{-4mm}
\label{fig:bbt_nmi_wcp_vs_nclust}
\end{figure*}

\begin{figure*}[t]
\begin{center}
\includegraphics[width=0.23\linewidth]{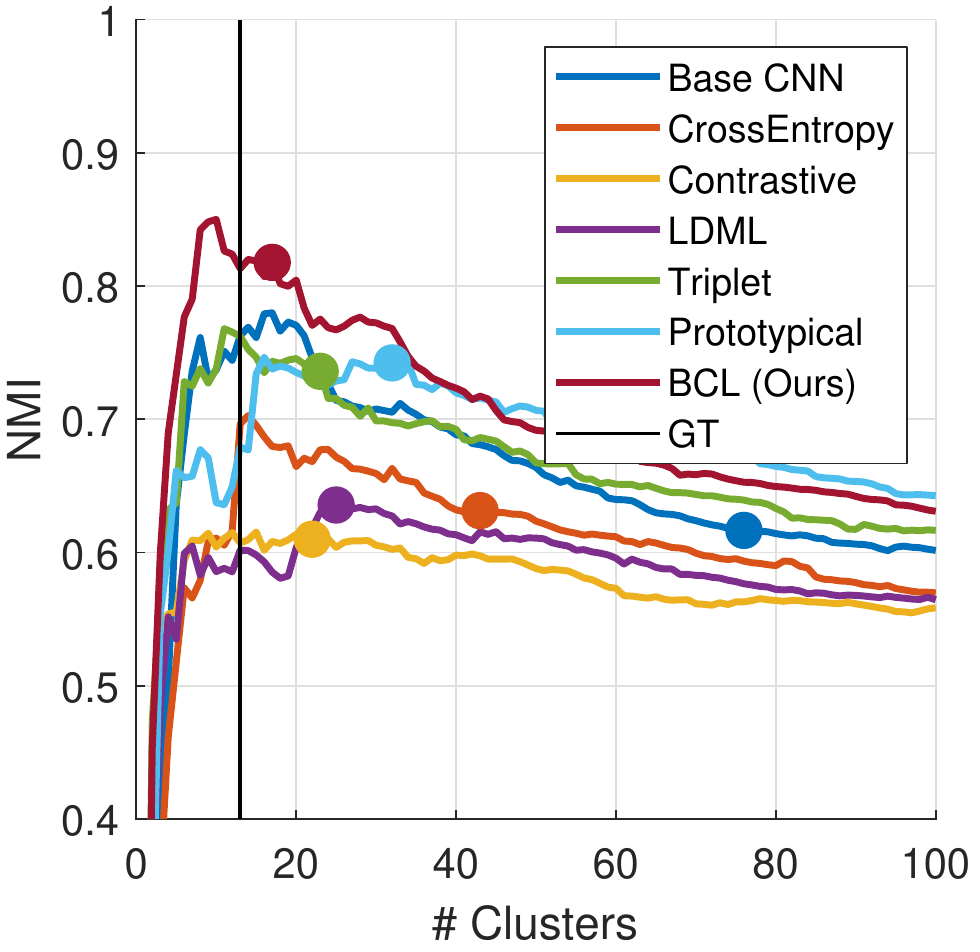}
\includegraphics[width=0.23\linewidth]{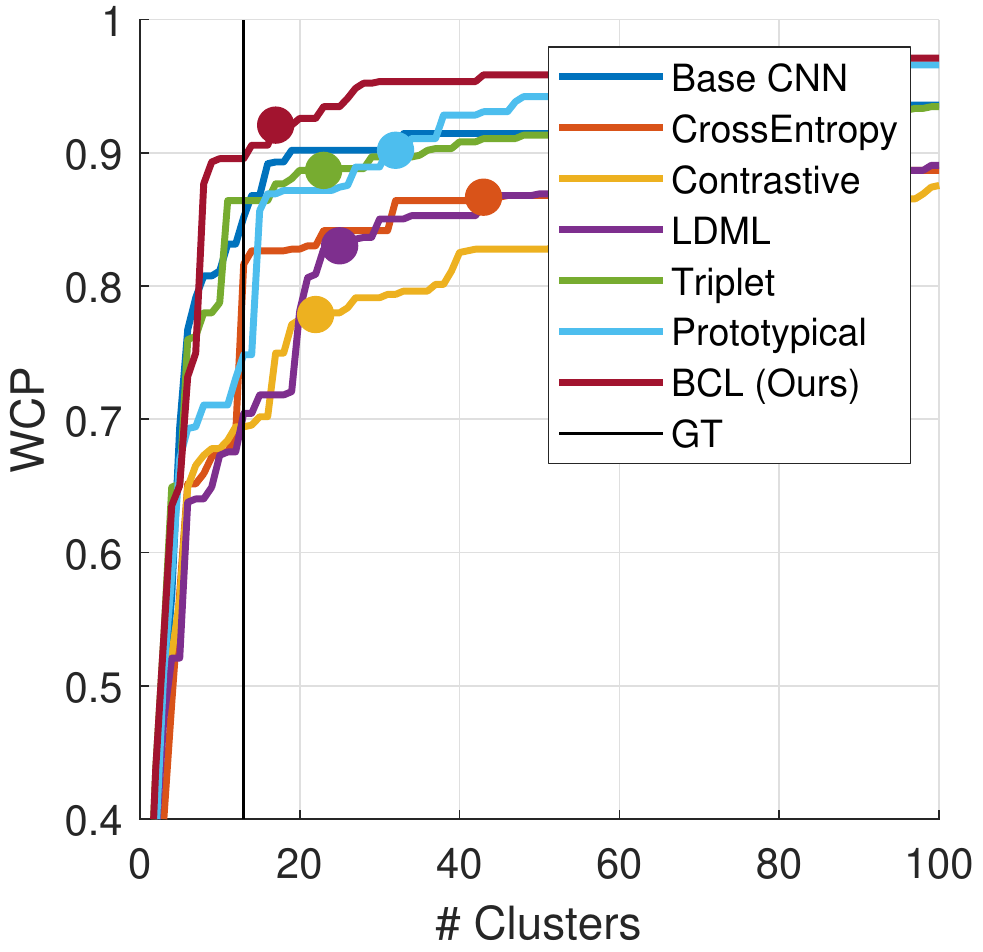} \quad\quad
\includegraphics[width=0.23\linewidth]{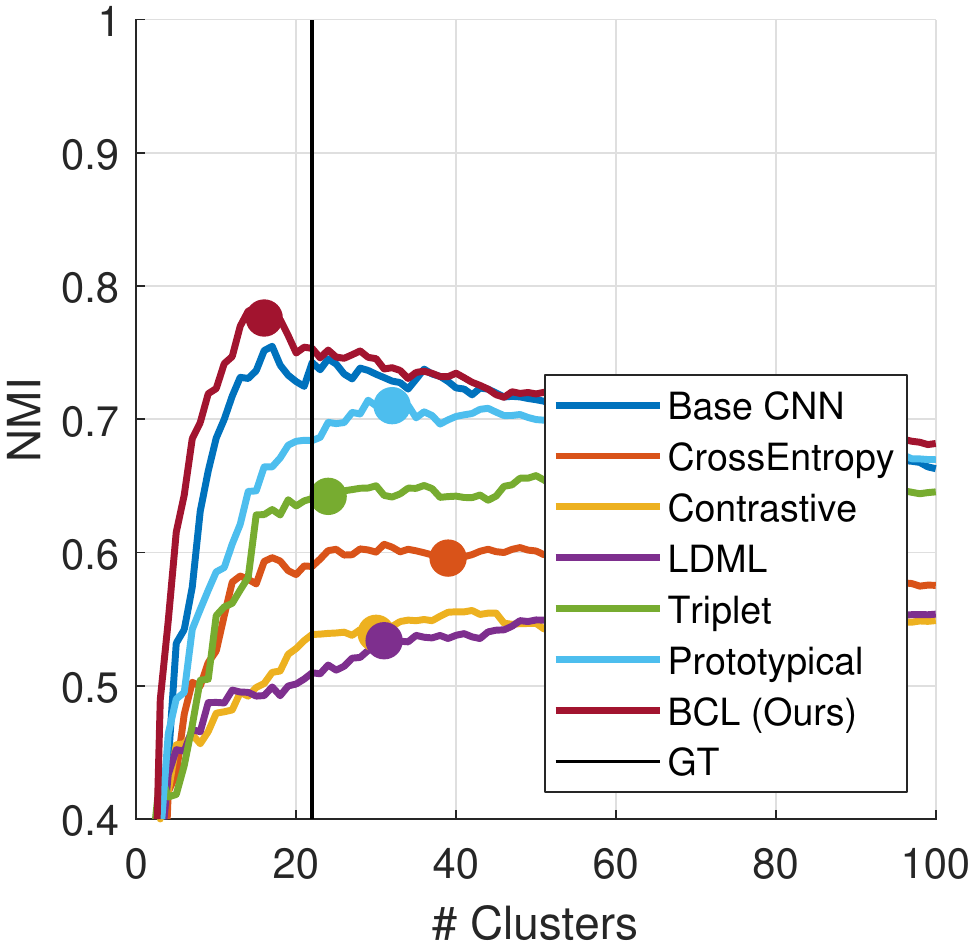}
\includegraphics[width=0.23\linewidth]{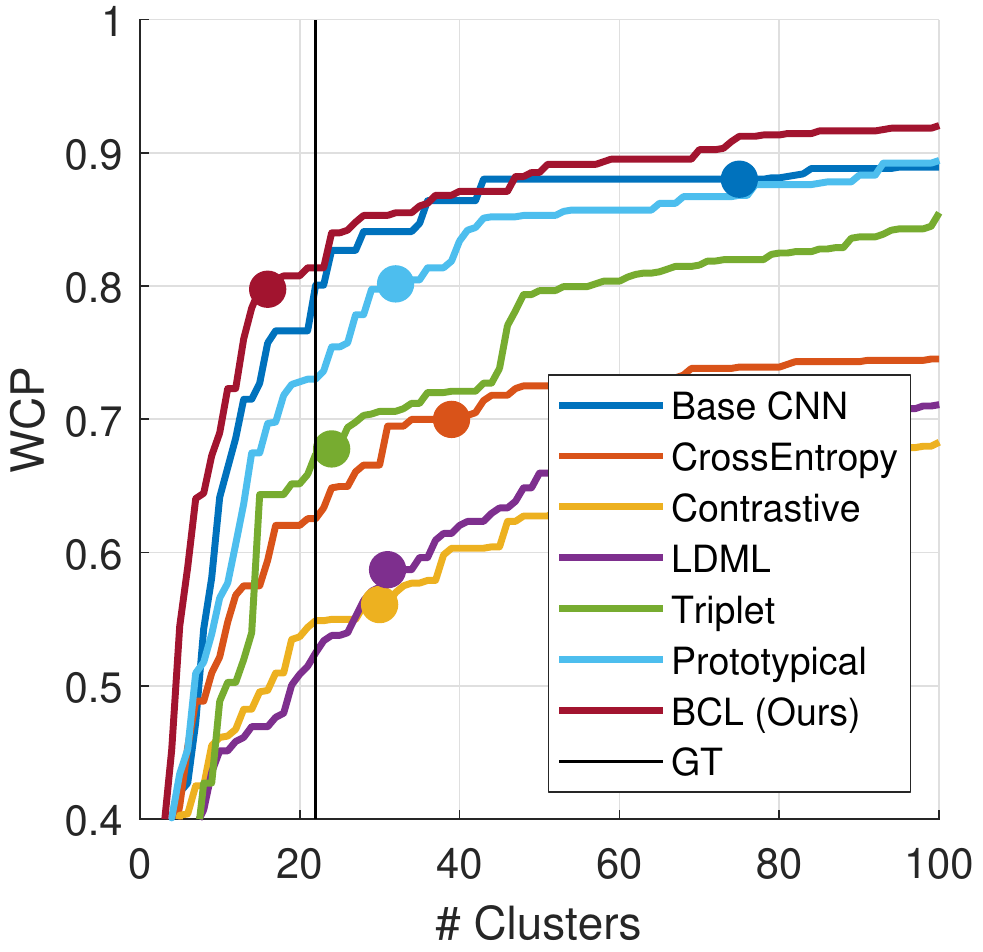} \\ \vspace{6mm}
\includegraphics[width=0.23\linewidth]{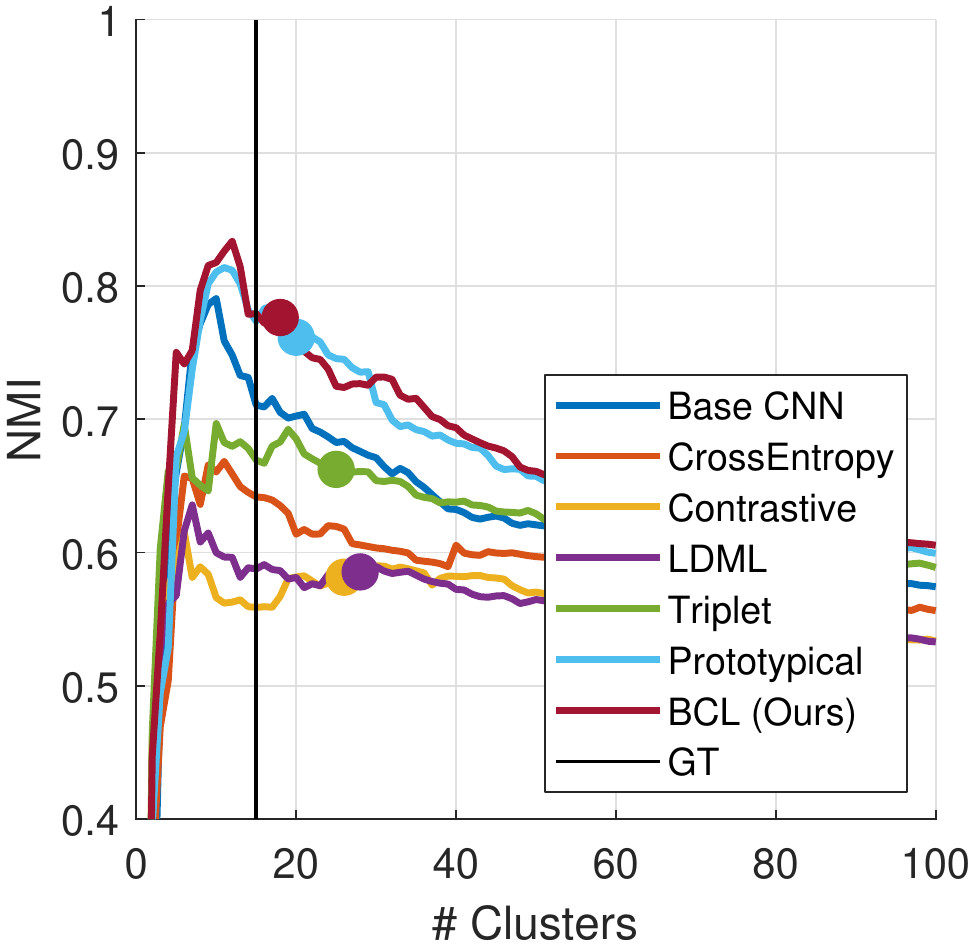}
\includegraphics[width=0.23\linewidth]{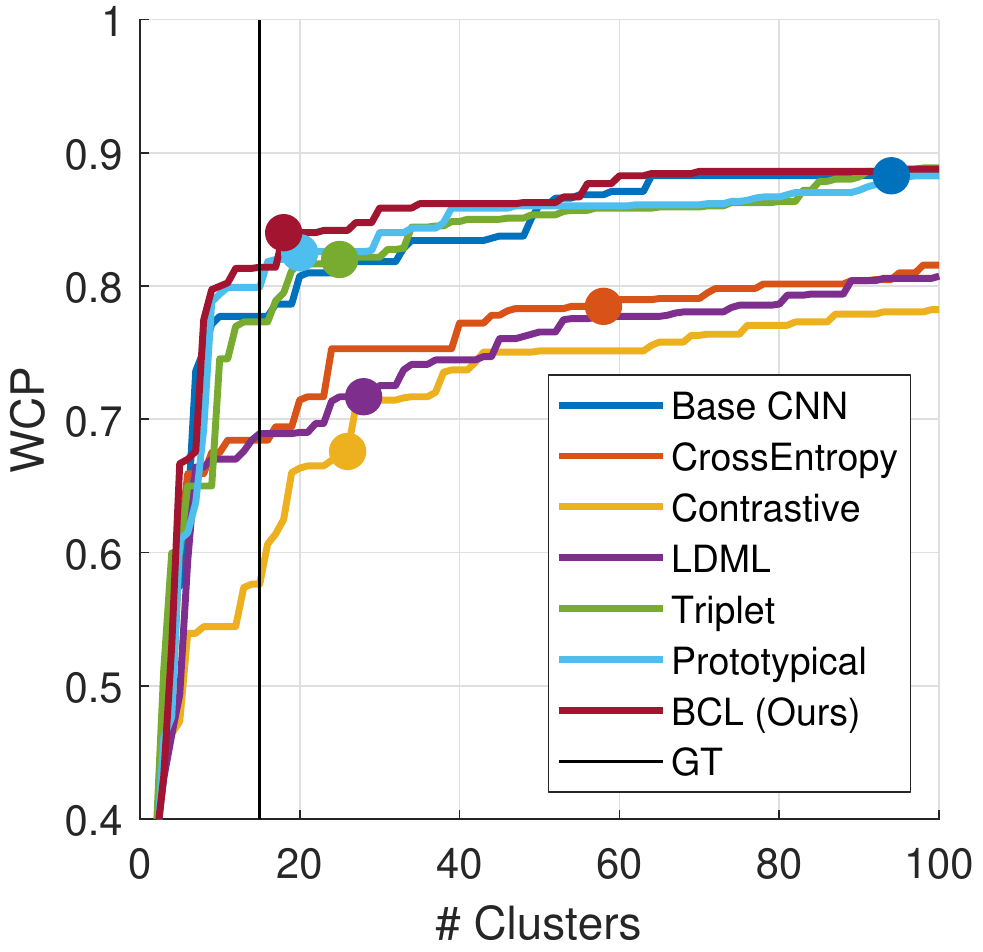} \quad\quad
\includegraphics[width=0.23\linewidth]{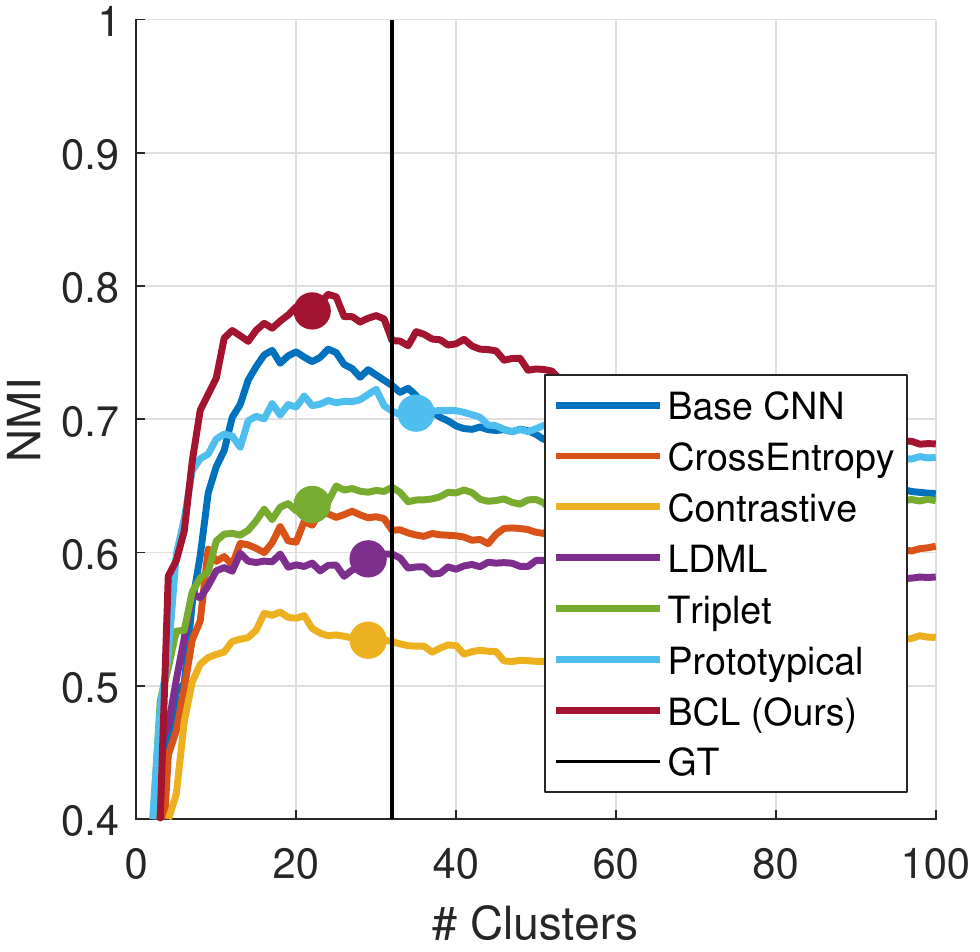}
\includegraphics[width=0.23\linewidth]{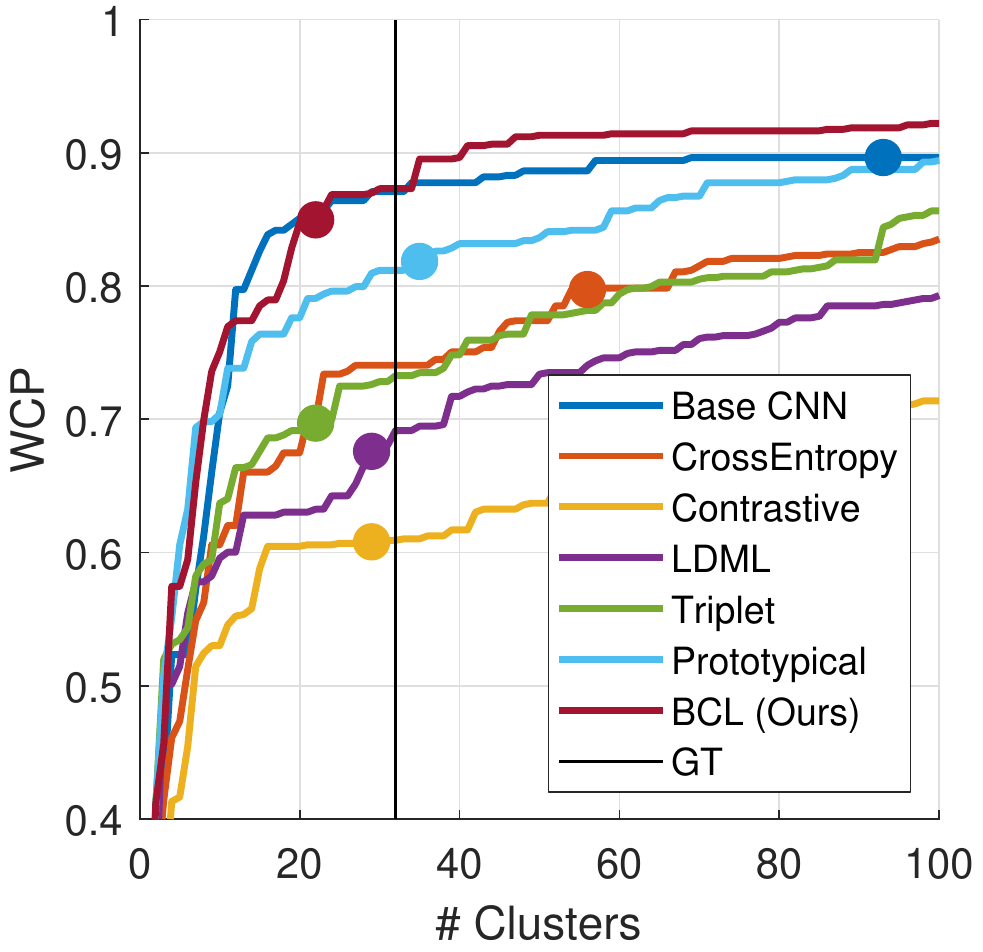} \\ \vspace{6mm}
\includegraphics[width=0.23\linewidth]{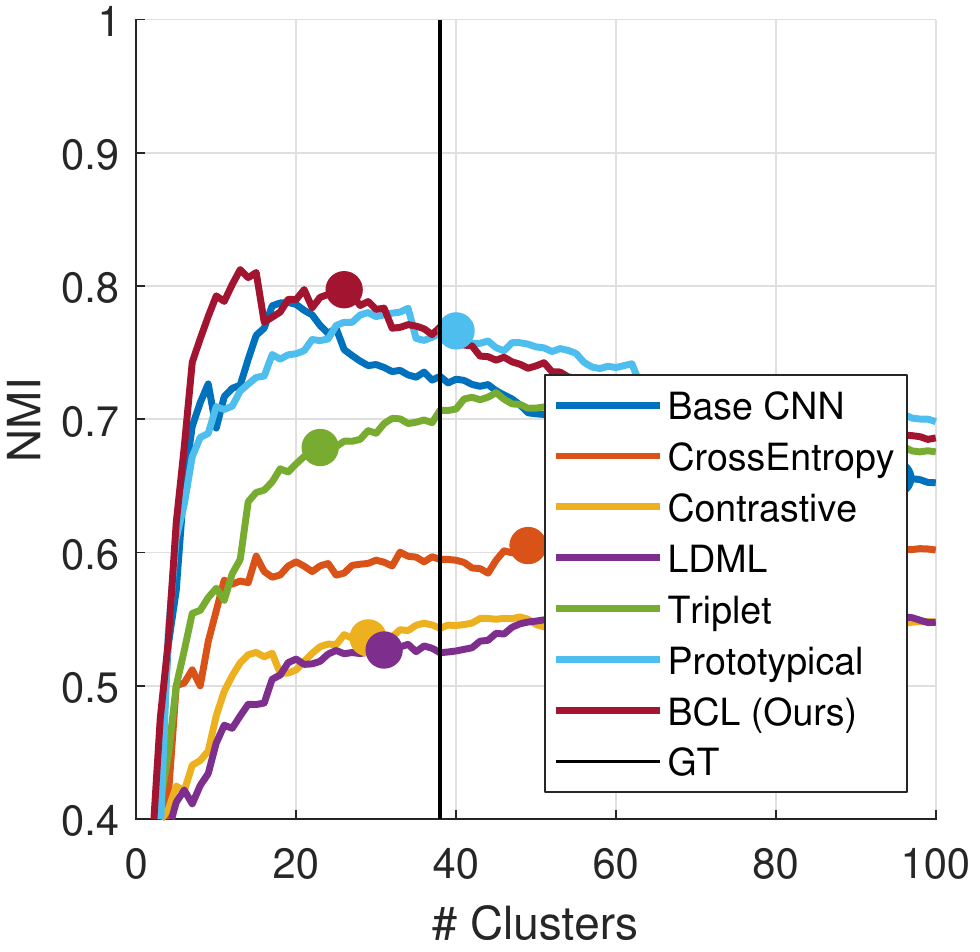}
\includegraphics[width=0.23\linewidth]{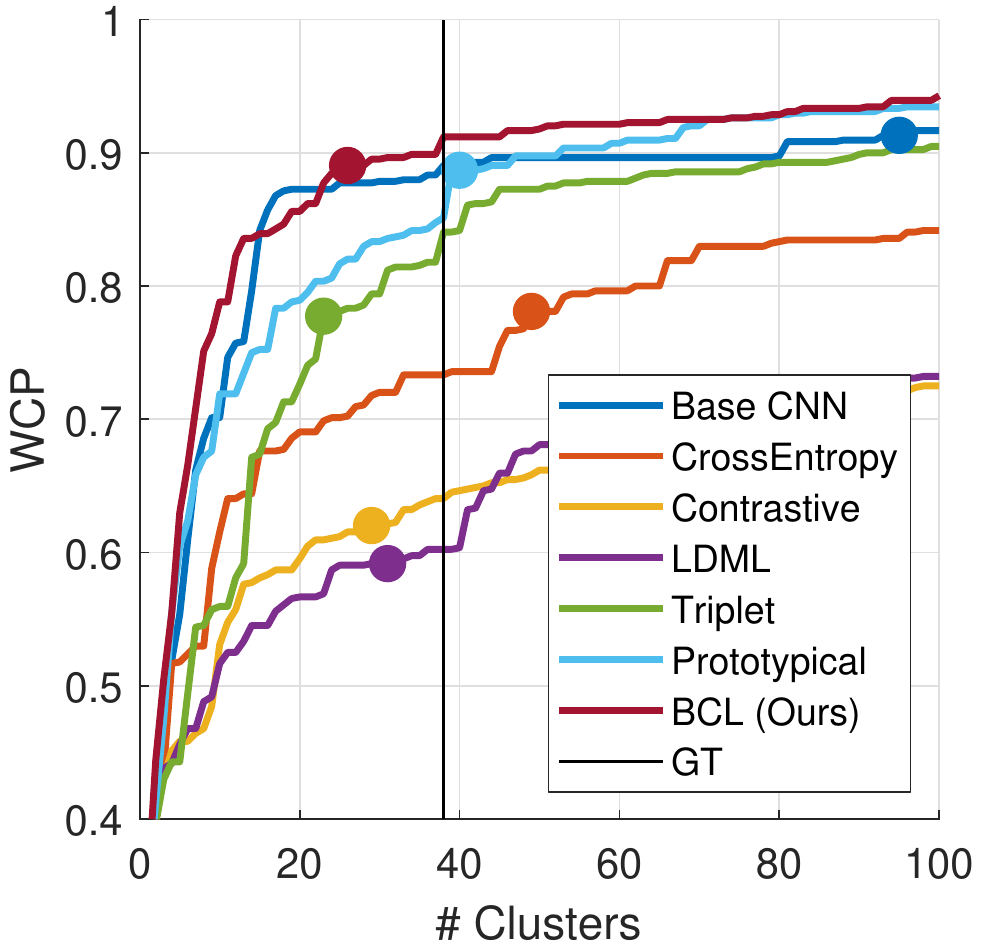} \quad\quad
\includegraphics[width=0.23\linewidth]{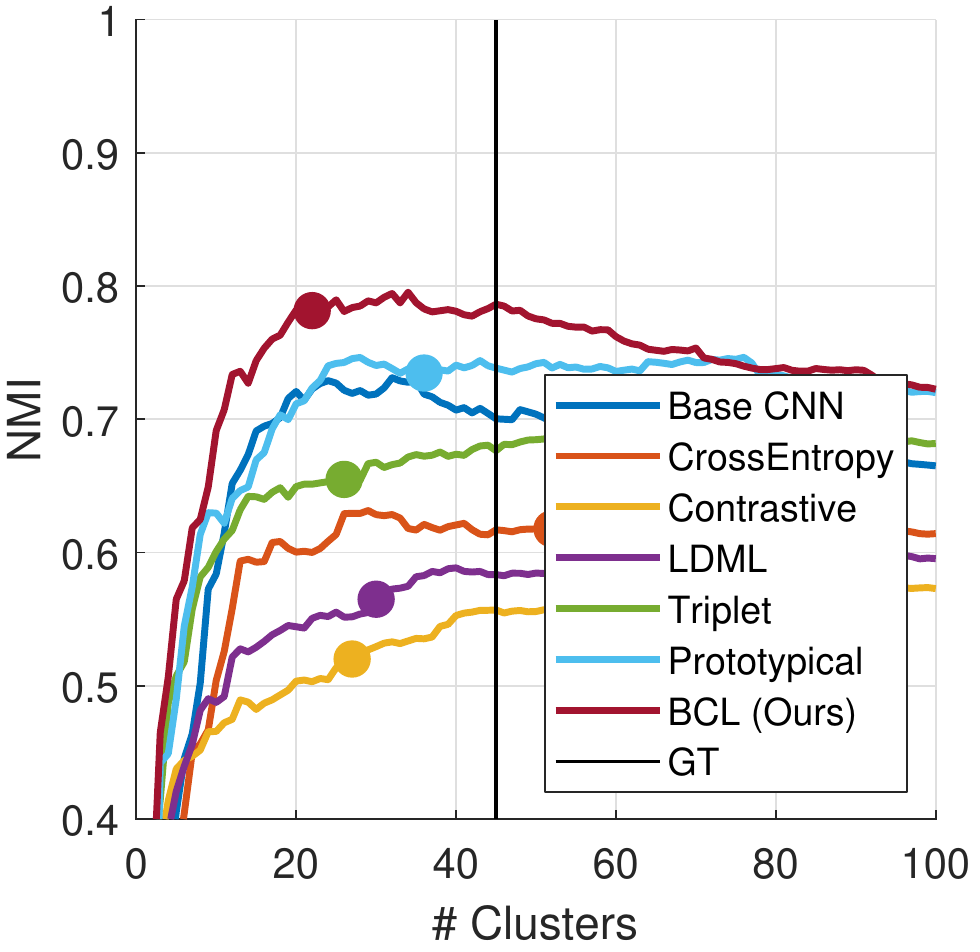}
\includegraphics[width=0.23\linewidth]{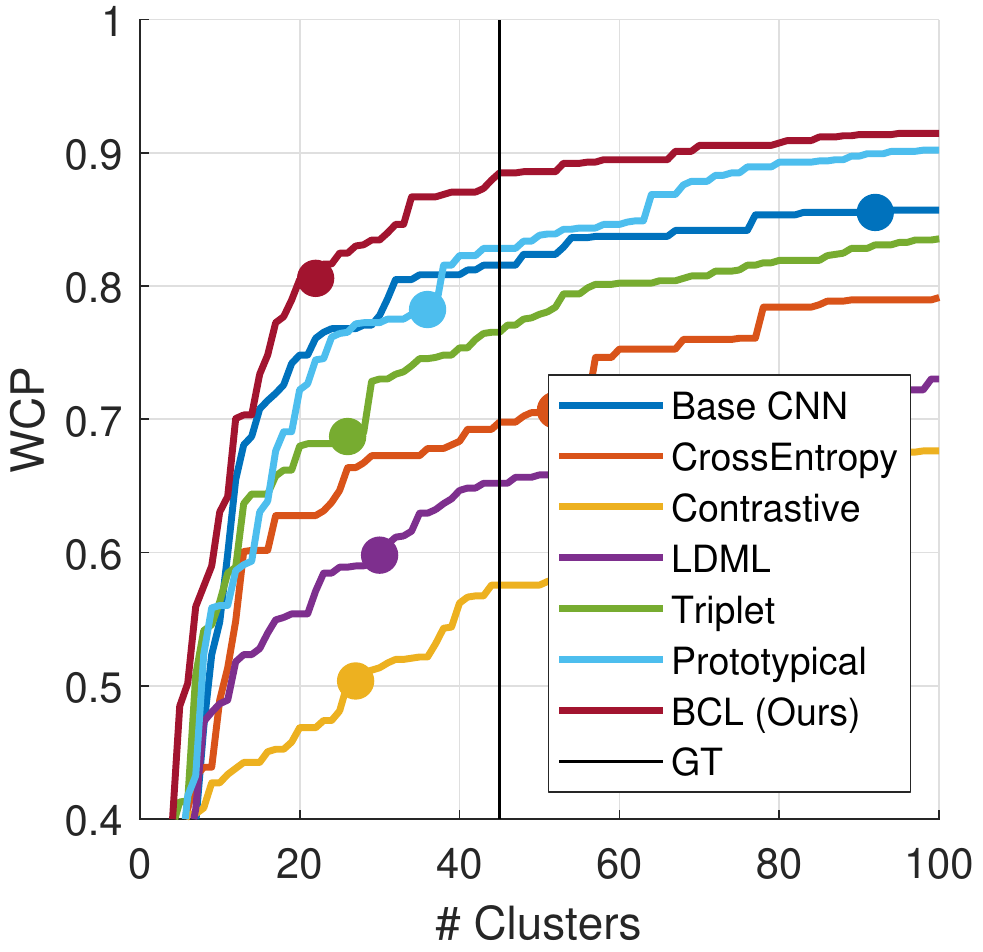}
\end{center}
\vspace{-2mm}
\caption{NMI and WCP vs. number of clusters for BUFFY-S5E1 to S5E6 (left to right, top to bottom).
Circles indicate operating points (\ie number of predicted clusters for the methods), our method uses the HAC threshold 4$b$, while all others are using the threshold tuned to give 66 clusters on the validation set.
Best seen in color.}
\vspace{-4mm}
\label{fig:buffy_nmi_wcp_vs_nclust}
\end{figure*}

\paragraph{NMI and WCP vs. number of clusters.}
We plot the NMI and WCP curves for all methods (as in Fig. \ref{fig:validation_losses} of the main paper) for each episode of BBT in Fig.~\ref{fig:bbt_nmi_wcp_vs_nclust} and BUFFY in Fig.~\ref{fig:buffy_nmi_wcp_vs_nclust}.
All results are before fine-tuning.
We wish to draw the reader to the following observations: \\
1. The threshold for prototypical loss is quite stable and is able to predict the number of clusters well (as was discussed in Table \ref{table:test_episodes} of the main paper). However, the {\bf purity is almost always lower than our method}, indicating that even though it makes more clusters, the formed clusters tend to be more heterogeneous (\ie contain more samples from different categories). \\
2. Our method shows higher NMI and WCP irrespective of the operating point in most episodes. \\
3. The base representations have very good performance curves. However, their operating points (chosen based on the validation threshold) are far from the optimal number of clusters.
Cross-entropy loss, especially when used with thousands of classes, seems to be effective at learning classification as well as clustering.

\begin{figure*}[p]
\begin{center}
\includegraphics[width=\linewidth]{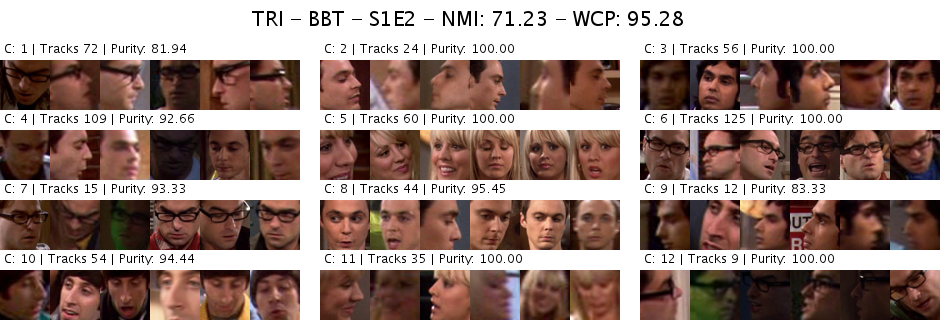} \vspace{4mm}
\includegraphics[width=\linewidth]{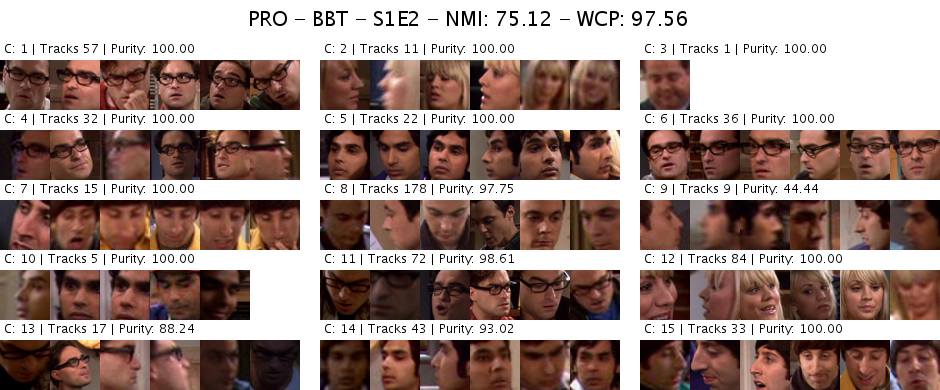} \vspace{4mm}
\includegraphics[width=\linewidth]{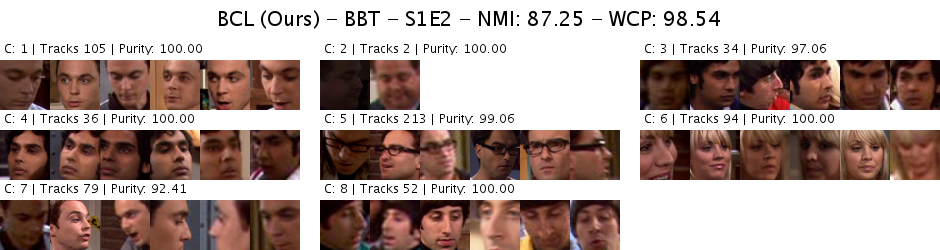}
\end{center}
\vspace{-2mm}
\caption{Clusters created by triplet loss (TRI, top), prototypical loss (PRO, middle), and BCL (bottom) on BBT-S1E2.
The correct number of clusters is 6.}
\vspace{-4mm}
\label{fig:bbt_clusters}
\end{figure*}

\begin{figure*}[p]
\begin{center}
\includegraphics[width=\linewidth]{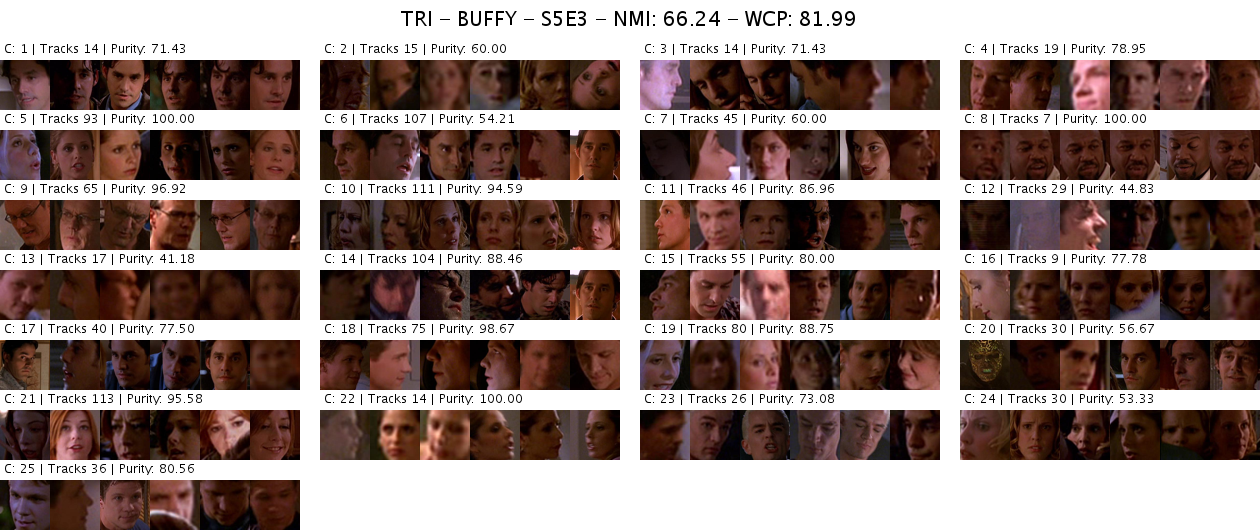} \vspace{4mm}
\includegraphics[width=\linewidth]{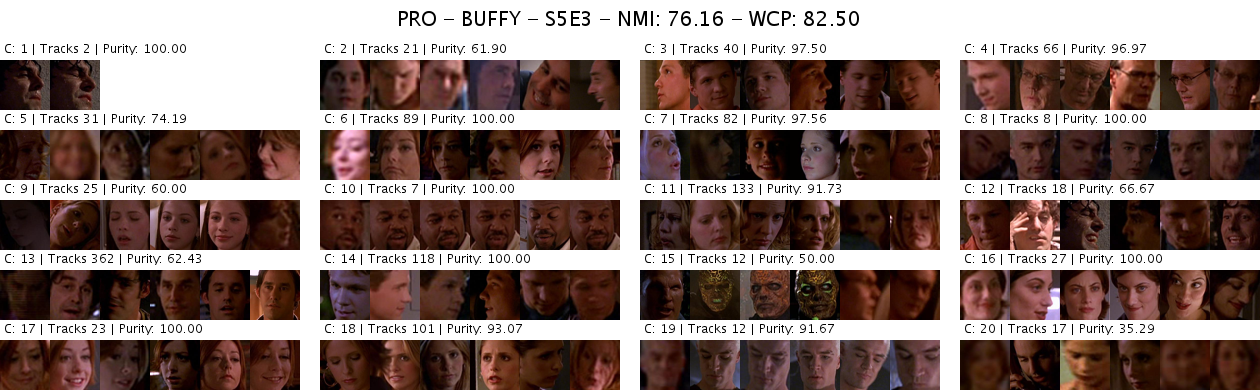} \vspace{4mm}
\includegraphics[width=\linewidth]{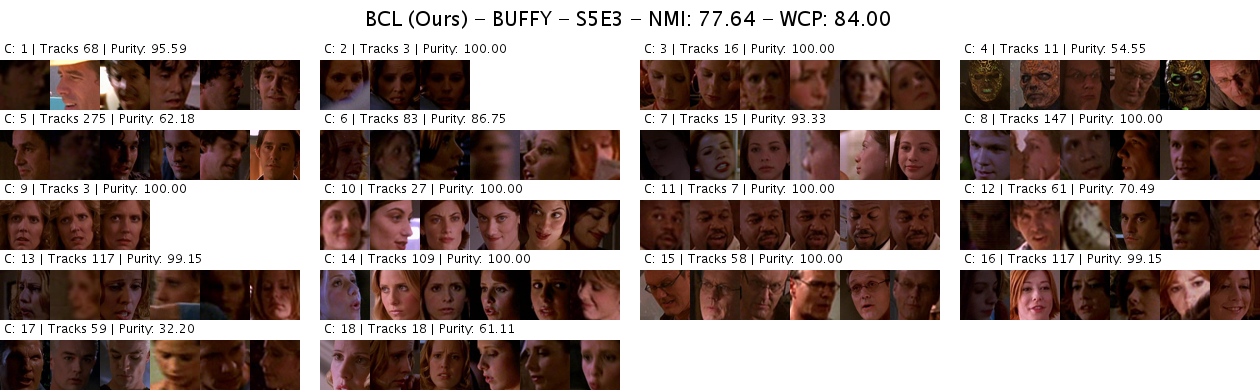}
\end{center}
\vspace{-2mm}
\caption{Clusters created by triplet loss (TRI, top), prototypical loss (PRO, middle), and BCL (bottom) on BUFFY-S5E3.
The correct number of clusters is 15.}
\vspace{-4mm}
\label{fig:buffy_clusters}
\end{figure*}

\paragraph{Qualitative visualization of clusters.}
Finally, we visualize the clusters created by Triplet loss (TRI), Prototypical Loss (PRO), and our method Ball Cluster Learning (BCL) on one episode of BBT (Fig.~\ref{fig:bbt_clusters}) and BUFFY (Fig.~\ref{fig:buffy_clusters}).
Each cluster is visualized by selecting 6 random face tracks (when available), and one face image per track.
All results are without fine-tuning.

These figures also throw light on the difficulty of our dataset that includes wide variations in illumination and pose.
Tracks, their labels and features, and our implementation of BCL is available at \url{https://github.com/makarandtapaswi/BallClustering_ICCV2019}.

In Fig.~\ref{fig:bbt_clusters}, BCL achieves close to the correct number of clusters, and separates the unknown character with just 2 tracks (C:2).
Both triplet and prototypical losses lead to over clustering.
\Eg~Leonard is split to C:1, C:6, C:7 and C:12 in triplet loss, and C:1, C:4, C:6, C:11, C:13 when using prototypical loss.

BUFFY-S5E3 (Fig.~\ref{fig:buffy_clusters}) is a unique episode in which one of the lead characters \emph{Xander} is duplicated due to a magic spell (the duplicate is played by the actor's identical twin).
Nevertheless, we see that BCL achieves reasonable performance, and is able to find minor characters (Joyce C:9, the building manager C:10), as well as isolate one of the background characters (C:11).

\end{document}